\documentclass[runningheads]{llncs}
\usepackage{booktabs}
\usepackage{graphicx}
\usepackage{amsmath}
\usepackage{amssymb}
\usepackage{subcaption}
\usepackage{orcidlink}
\usepackage[subtle]{savetrees}
\usepackage{cleveref}
\Crefformat{figure}{#2Fig.~#1#3}

\usepackage[utf8]{inputenc}
\usepackage{pgfplots}
\DeclareUnicodeCharacter{2212}{−}
\usepgfplotslibrary{groupplots,dateplot}
\usetikzlibrary{patterns,shapes.arrows}
\pgfplotsset{compat=newest}
\pgfplotsset{compat=newest}
\pgfplotsset{legend image post style={black},legend style = {text=black}}
\usepackage{hyperref}

\begin{document}
\title{Drawing the Line:
Deep Segmentation for Extracting Art from Ancient Etruscan Mirrors\thanks{This project has received funding from the Austrian Science Fund/Österreichischer Wissenschaftsfonds (FWF) under grant agreement No. P~33721.}}
\titlerunning{Extracting Art from Ancient Etruscan Mirrors}
%
\author{Rafael Sterzinger\orcidlink{0009-0001-0029-8463} \and
Simon Brenner\orcidlink{0000-0001-6909-7099} \and
Robert Sablatnig\orcidlink{0000-0003-4195-1593}}
\authorrunning{R. Sterzinger et al.}
%
\institute{Computer Vision Lab, TU Wien, Vienna, AUT\\
\email{\{firstname.lastname\}@tuwien.ac.at}}
\maketitle              
\begin{abstract}
Etruscan mirrors constitute a significant category within Etruscan art and, therefore, undergo systematic examinations to obtain insights into ancient times.
A crucial aspect of their analysis involves the labor-intensive task of manually tracing engravings from the backside.
Additionally, this task is inherently challenging due to the damage these mirrors have sustained, introducing subjectivity into the process.
We address these challenges by automating the process through photometric-stereo scanning in conjunction with deep segmentation networks which, however, requires effective usage of the limited data at hand.
We accomplish this by incorporating predictions on a per-patch level, and various data augmentations, as well as exploring self-supervised learning.
Compared to our baseline, we improve predictive performance w.r.t.\ the pseudo-F-Measure by around~16\%.
When assessing performance on complete mirrors against a human baseline, our approach yields quantitative similar performance to a human annotator and significantly outperforms existing binarization methods.
With our proposed methodology, we streamline the annotation process, enhance its objectivity, and reduce overall workload, offering a valuable contribution to the examination of these historical artifacts and other non-traditional documents.
\keywords{Binarization \and Photometric Stereo \and Image Segmentation \and Limited Data \and Etruscan Art \and Cultural Heritage}
\end{abstract}

\section{Introduction}

\begin{figure}[t]
   \centering
   \includegraphics[width=0.60\textwidth]{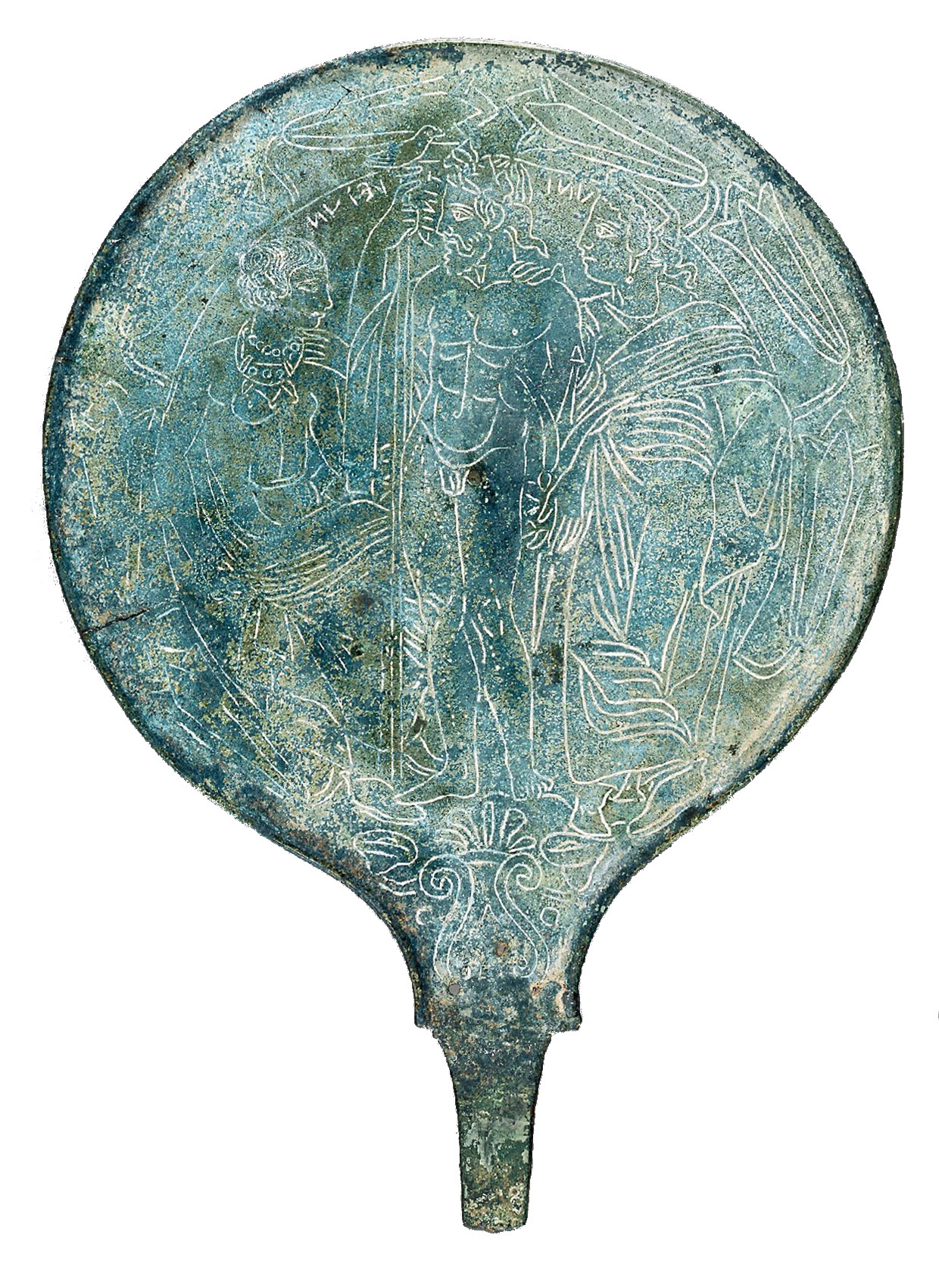}
   \caption[A typical Etruscan mirror: fine drawings adorn its backside.]{A typical Etruscan mirror: fine drawings adorn its backside.\footnotemark}
   \label{fig:example}
\end{figure}

Comprising more than 3,000 identified specimens, Etruscan hand mirrors represent one of the most extensive categories in Etruscan art.
These ancient artworks, which date from the second half of the sixth century to the first century B.C.E., are predominantly made from bronze.
On the front, they feature a highly polished surface acting as a mirror, whereas the back side typically depicts engraved and/or chased linear figurative illustrations of Greek mythology which may be supplemented by inscriptions in Etruscan~\cite{sindy_2023}.
An exemplary mirror showcasing such drawings is depicted in \Cref{fig:example}.

These mirrors have been systematically examined since the 1980s to better understand the Etruscan culture, with findings published in the scholarly series \textit{Corpus Speculorum Etruscorum}~(e.g., \cite{swaddling1993corpus,wiman2018corpus}).
One main component of their examination is the labor- and cost-intensive task of tracing the linear figurative artworks typically featured on their backside.
Although performed by experts,~i.e., archaeologists specialized in Etruscan culture, this task is prone to subjectivity: over time, these mirrors experienced erosion/corrosion, which increases the difficulty of distinguishing between artwork, damage, and background.\footnotetext[1]{\textcopyright KHM-Museumsverband (Kunsthistorisches Museum Vienna)}

Based on these challenges, this work presents the first approach for automated segmentation of Etruscan hand mirrors to enhance annotation quality and objectivity while reducing the workload of manual tracing.
For this, we collect high-resolution 3D surface models via photometric stereo and train a deep segmentation network to recognize intentional lines over scratches.
Our approach is grounded in a thorough ablation study where we evaluate various surface modalities and explore different techniques to address the scarcity of data and annotations within this domain, including performing predictions on a per-patch level, incorporating various data augmentations, and self-supervised learning, as well as examine different architectural components, such as encoders, decoders, and loss functions.
Compared to an initial baseline, our modifications improve performance w.r.t.\ the pseudo-F-Measure by around 16\%. 
Moreover, we juxtapose our method with a human baseline to underscore its efficacy as it achieves results on par with a human annotator.

Summarizing our contributions: our methodology streamlines the overall annotation process of Etruscan hand mirrors; It not only expedites the process of manual tracing but also offers a supplementary perspective, contributing to increased quality and objectivity.
Finally, we provide public access to both the code and data utilized in this work (see \href{https://github.com/RafaelSterzinger/etmira-segmentation}{github.com/RafaelSterzinger/etmira-segmentation}) to promote transparency and reproducibility.

\section{Related Work}
Photometric Stereo (PS) is a powerful tool to capture unique insights into an object's surface geometry.
McGunnigle et al.~\cite{mcgunnigle2003resolving} demonstrate its usefulness in segmentation by extracting handwriting on paper based on depth profiles.
Similarly, Landström et al.~\cite{LandstromThurleyJonsson2013} have applied this technique to detect cracks in steel surfaces, showing its versatile application.
Moreover, the work by Zhang et al.~\cite{ZhangHansenSmithSmithGrieve2018}, where they compute shape index and curvedness from PS for leaf venation extraction, or by Tao et al.~\cite{TaoGongWangLuoQiuHuang2022}, in which they detect air voids in concrete surfaces, further exemplify its broad applicability.
Its ability to distinguish subtle surface variations makes it particularly suitable for analyzing complex patterns such as those found on Etruscan hand mirrors.
However, traditional segmentation methods might fall short due to damage these mirrors experienced, increasing the difficulty of distinguishing between fore- and background.

In the field of image segmentation, the state of the art is dominated by deep learning architectures such as the UNet~\cite{ronneberger_u-net_2015}, DeepLabV3++~\cite{chen_encoder-decoder_2018}, Pyramid Scene Parsing Network~\cite{zhao_pyramid_2017}, Pyramid Attention Network~\cite{li_pyramid_2018}, or the Feature Pyramid Network~\cite{lin_feature_2017}.
A driving force for such innovations is, for instance, the medical sector, which demands high-detailed segmentations as well.
One example is the segmentation of vascular structures found in the retina (e.g., see \cite{li_iternet_2019}), which is key in determining retinal diseases.
In this context, Kamran et al.~\cite{kamran_rv-gan_2021} proposed a multiscale, GAN-based segmentation system, designed to reduce loss in fidelity.
More recent works propose transformer-based~\cite{valanarasu_medical_2021} as well as diffusion-based architectures~\cite{wu_medsegdiff_2023} for brain tumor, gland, or nuclei segmentation.

Saiz et al.~\cite{SaizBarandiaranArbelaizGrana2022,SaizAlfaroBarandiaranGrana2021} showcased the effectiveness of combining PS scanning with deep segmentation networks.
With this paradigm, they perform defect detection in steel component manufacturing, paving the way for future applications where intricate and detailed segmentation is vital. 
Lastly, to give an example in the context of cultural heritage, Redon et al.~\cite{redon_3d_nodate} present a deep learning-based reconstruction pipeline that transforms RGB images into high-resolution normal maps to digitize Bayeux tapestry.
Additionally, they couple the reconstruction with segmentation, enabling cost-effective and quick production of 3D-printed bas-reliefs to enhance tactile exploration for visually impaired individuals.
Such examples showcase the potential efficacy of PS when combined with deep learning to segment figurative drawings such as the ones found on Etruscan mirrors.

\section{Methodology}
In the following, we describe the setup for data acquisition, the captured dataset, and outline our pipeline in detail, focusing on preprocessing and the subsequent segmentation.
For the latter, we summarize our explored augmentation/self-supervised learning techniques as well as point out training and inference specifics.

\begin{figure}[t]
    \begin{center}
 	\begin{subfigure}{0.44\textwidth}
		\includegraphics[width=\textwidth]{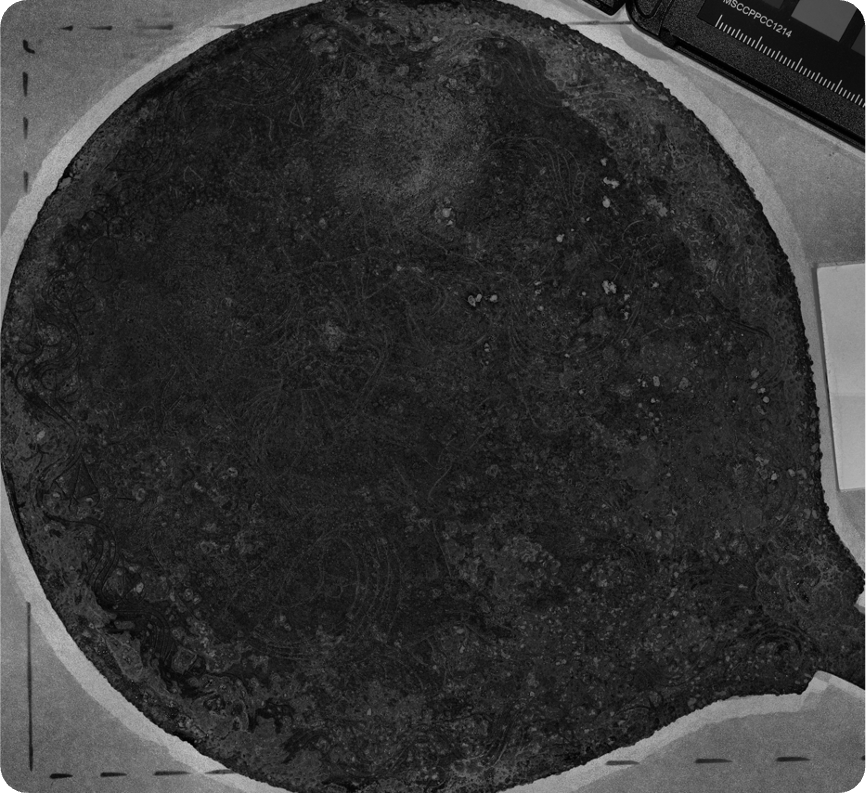}
		\caption{Albedo Map}
	\end{subfigure}
    \hspace{0.04\textwidth}
    \begin{subfigure}{0.44\textwidth}
	        \includegraphics[width=\textwidth]{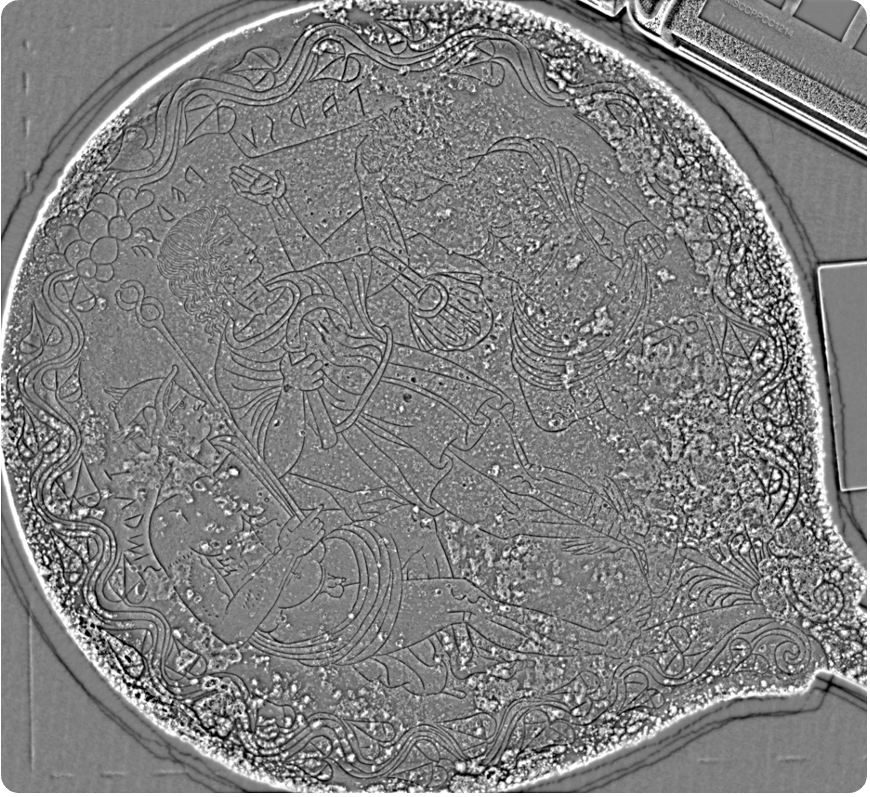}
	        \caption{High-Pass Filtered Depth Map}
    \end{subfigure}
    \end{center}
    \begin{center}
	\begin{subfigure}{0.44\textwidth}
		\includegraphics[width=\textwidth]{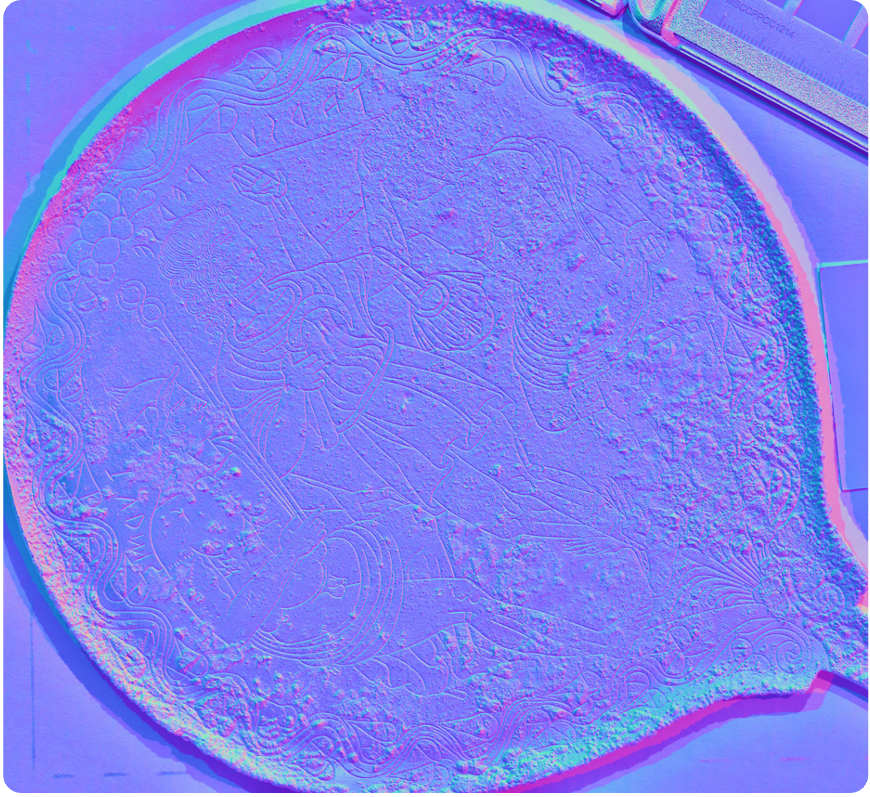}
		\caption{Normal Map}
	\end{subfigure}
    \hspace{0.04\textwidth}
	\begin{subfigure}{0.44\textwidth}
	        \includegraphics[width=\textwidth]{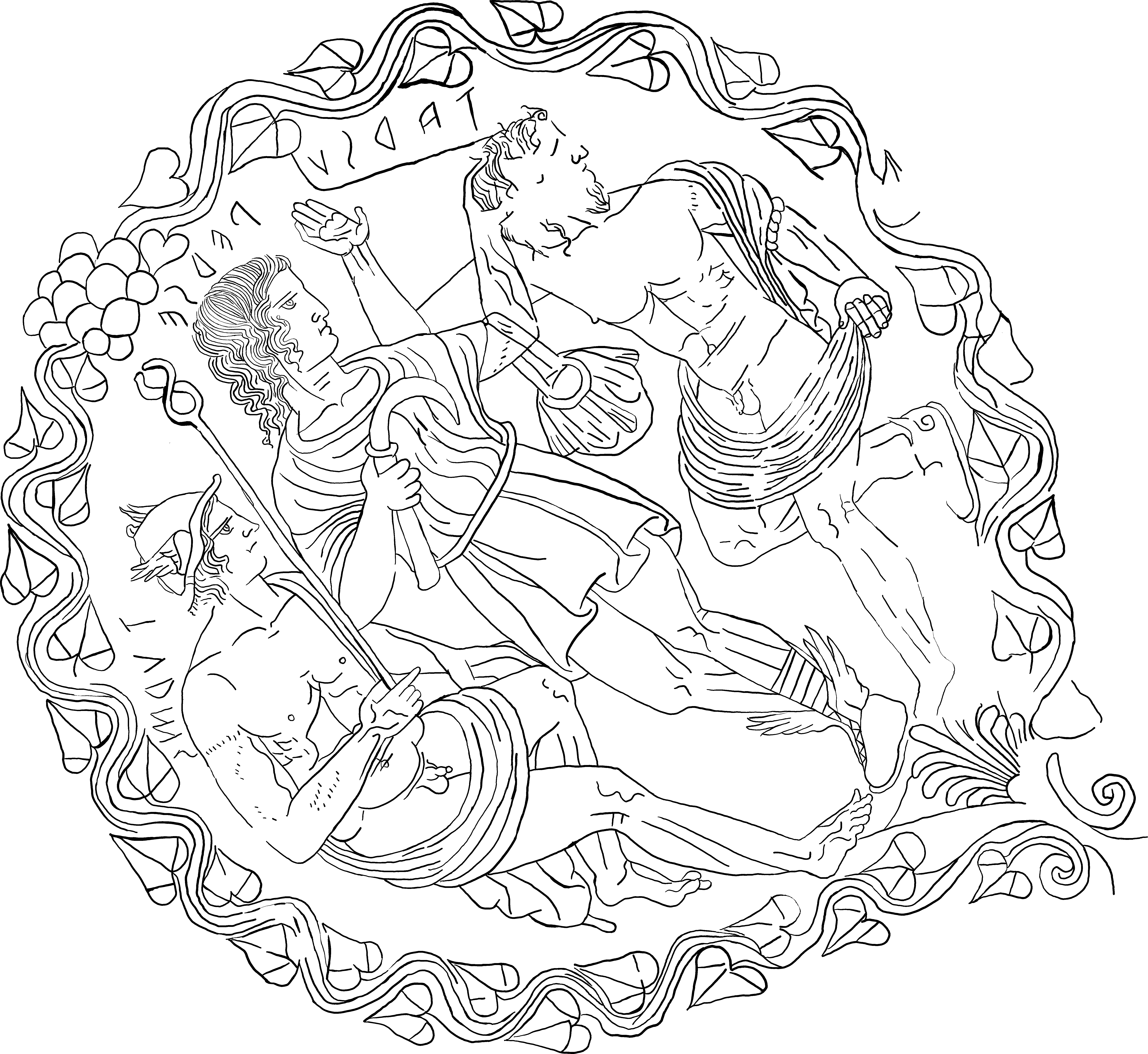}
	        \caption{Ground Truth Mask}
    \end{subfigure}
    \end{center}
    \caption{Exemplary albedo, depth, normal map, and corresponding ground truth.}
    \label{fig:input}
\end{figure}

\subsection{Data Acquisition}\label{sec:data_acquition}

We employ PS, a 3D reconstruction technique introduced by Woodham \cite{woodham1980photometric} in~1980. 
Compared to other 3D acquisition methods, e.g., structured-light scanning or photogrammetry, it is especially adept at capturing local surface details~\cite{herbort_introduction_2011}.
PS accomplishes this by capturing a series of images with consistent camera settings but variable lighting conditions to then, for each pixel, determine surface orientation, distance, and illumination-independent brightness.
These properties can be encoded across the entire image to \textit{normal, depth} (computed from the normal map via integration \cite{55103}), and \textit{albedo maps}, illustrated in \Cref{fig:input}. 
 
For recording the images, we employ a PhaseOne IQ260 Achromatic camera with an $8,964\times6,716$ pixel medium-format sensor and a Schneider-Kreuznach 120 mm LS Macro lens.
Using this setup, we obtained a surface resolution of~38.5~ppm ($\sim$978 dpi) for most of our mirrors.
However, as imaging distance changed between locations, this value is not consistent throughout the images, ranging from 29.8 ppm to 58.3 ppm.
For more information, we refer the reader to Brenner et al.~\cite{brenner_revealing_2023} from where we derived our setup.

\subsection{Dataset}
Our dataset includes a wide variety of Etruscan mirrors preserved in public collections in Austria.
Its core comprises 53 specimens located at the Kunsthistorischen Museum (KHM) Vienna and 6 additional ones scattered throughout Austria.
We acquired annotations for 19 out of these 59 mirrors, whereby 19 backsides and 10 fronts were annotated, leading to 29 examples in total.
Note that engravings mainly decorate the backside where they do not interfere with reflectance.
Sometimes, however, they can also be found on the front near the handle or around the border, though with less dense drawings.

We split these labels into training, validation, and test sets, however, given their limited amount, their varying density of engravings per mirror, and their condition, this step is not straightforward: first, mirrors with dense engravings offer a stronger learning signal and are hence more preferable for the training, second, using mirrors of different condition for validation and testing, will cause discrepancies in performance, rendering model selection difficult.
Therefore, we select the backsides of three mirrors of different conditions and densities in engravings for evaluation, one mirror from Wels and two from the KHM Vienna.
We create non-overlapping patches of size $512\times512$ pixels, shuffle them, and split them in half to yield our validation and test set, to create splits with similar underlying distributions.
Out of the remaining 26 examples, we remove one, representing an outlier, due to its different art style (points instead of lines), leaving us with 25 annotated examples for training. 

\begin{figure}[t]
	\begin{center}
	\begin{subfigure}{0.44\textwidth}
		\includegraphics[width=\textwidth]{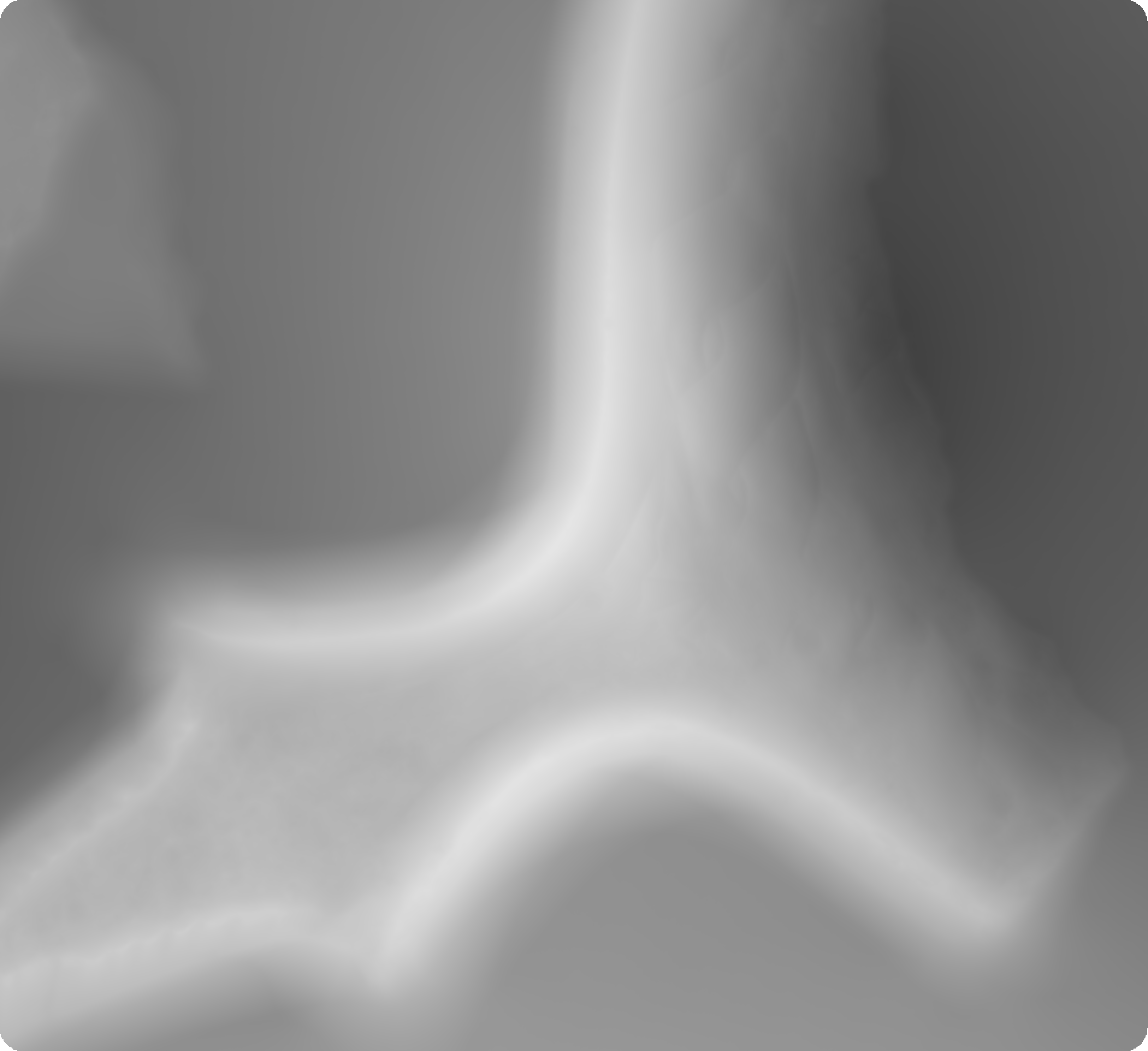}
        \caption{Raw Image}
	\end{subfigure}
    \hspace{0.04\textwidth}
	\begin{subfigure}{0.44\textwidth}
		\includegraphics[width=\textwidth]{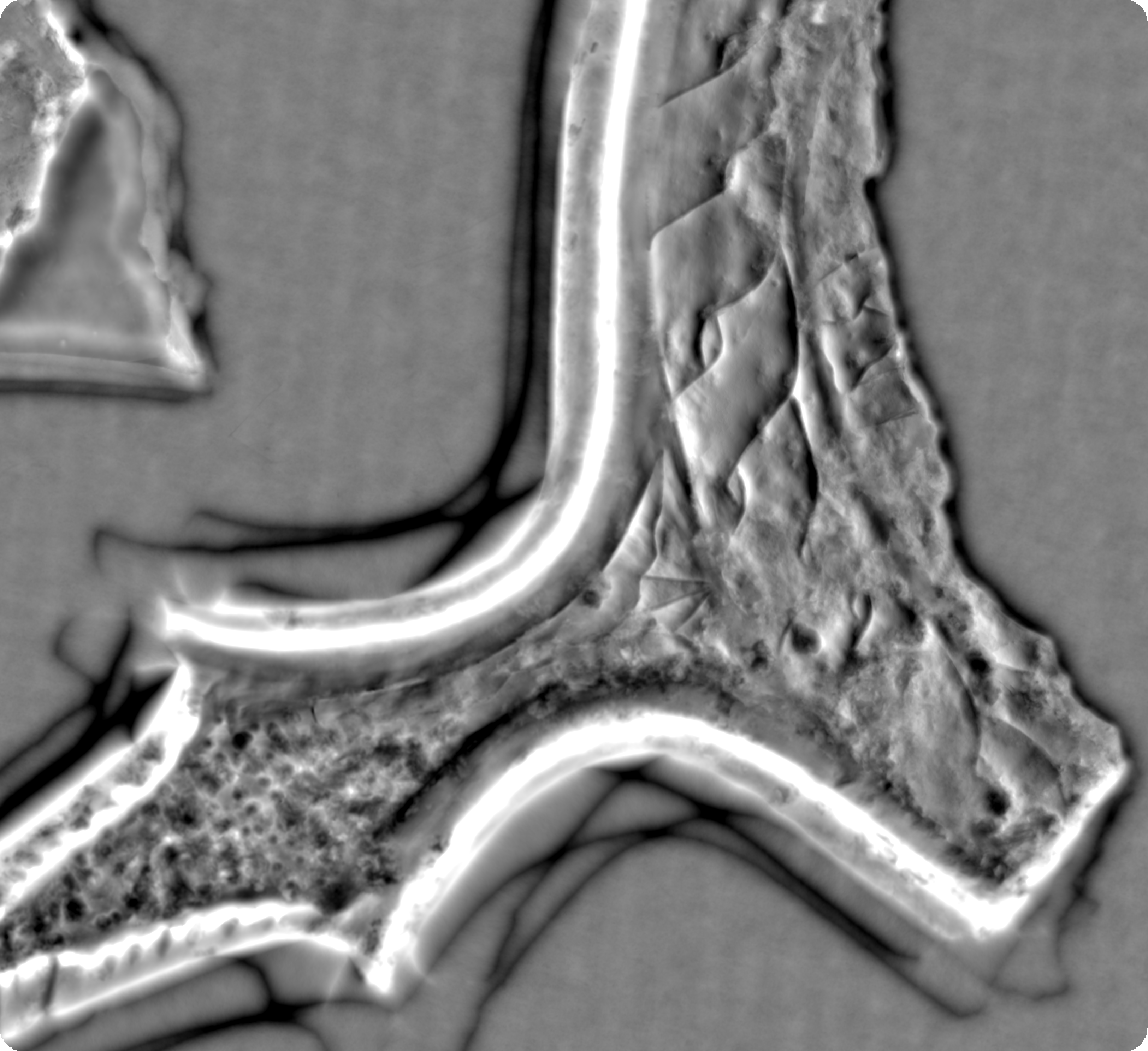}
        \caption{High-Pass Filtered Image}
	\end{subfigure}
    \end{center}
    \caption{Comparing the effect of preprocessing on the depth modality, i.e., employing a high-pass filter to remove low frequencies and to make lines more visible.}
    \label{fig:preprocessing}
\end{figure}

\subsection{Preprocessing}

With three modalities at hand, albedo, depth, and normal, it is unclear which ones obtain the best predictive performance.
In order to allow for a fair performance evaluation, preprocessing is crucial as it directly influences the result of subsequent tasks.
For instance, employing the depth modality and not performing any preprocessing renders the segmentation of fine artistic lines impossible; its effect is illustrated in~\Cref{fig:preprocessing}.
Here, we filter low frequencies by subtracting a Gaussian-filtered version of the depth map.
Afterward, we cap the depth values to~$\mu - 3\sigma$ as the minimum and~$\mu + 3\sigma$ as the maximum intensity, where~$\mu$ is the mean and~$\sigma$ is the standard deviation.
Regarding the other two modalities,~i.e., normal and albedo, we standardize the normal vectors to be of unit length, whereas we keep the latter modality unmodified.

In addition to that, we generate segmentation masks to capture the location of a mirror within a shot using the Segment Anything Model~(SAM)~\cite{kirillov_segment_2023}.
Although these masks are not necessarily required, we employ them for Cross-Pseudo Supervision (CPS)~\cite{chen_semi-supervised_2021} as well as during inference to mask out regions that impossibly contain engravings.
Lastly, using the acquired masks, we ignore non-mirror parts for the calculation of per-channel means and standard deviations which are later on used to normalize the input.

As mentioned, to alleviate our lack of annotations, we perform inference on a per-patch level:
For validation and testing, we extract non-overlapping, quadratic patches of size~$512\times512$ pixels.
Regarding our training data, we pad four pixels, since $6720 \equiv 0 \pmod{2240}$, to the original resolution ($8,964\times6,716$~pixels) to extract 25 overlapping tiles of size $2,988\times2,240$~pixels 
using a stride of half the size; tiles, containing no annotation, are discarded.
Later, during training, to create varied training data for one epoch, we extract ten patches per tile, all of the same size as our validation/test patches.
Finally, all patches are resized to a height and width of $256\times256$ pixels to reduce model complexity.

\begin{figure*}[t]
	\centering
	\begin{subfigure}{0.48\textwidth}
		\includegraphics[width=\textwidth]{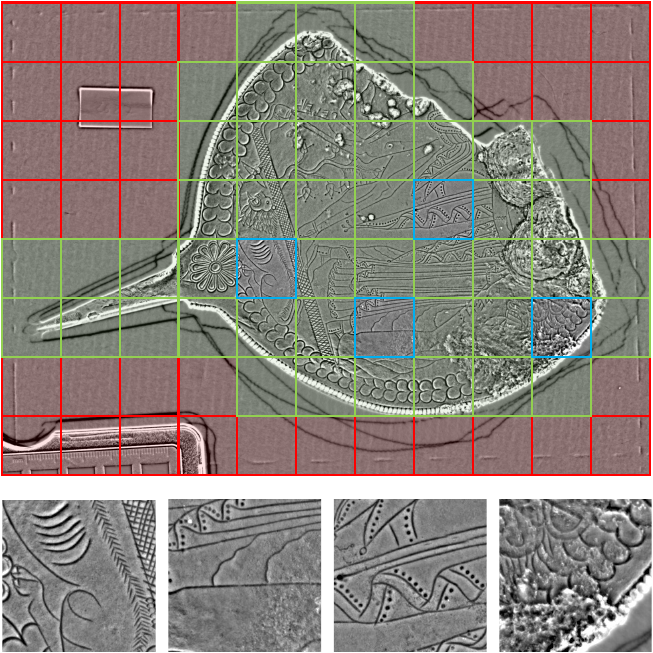}
		\caption{Input}
	\end{subfigure}
    \hfill
	\begin{subfigure}{0.48\textwidth}
		\includegraphics[width=\textwidth]{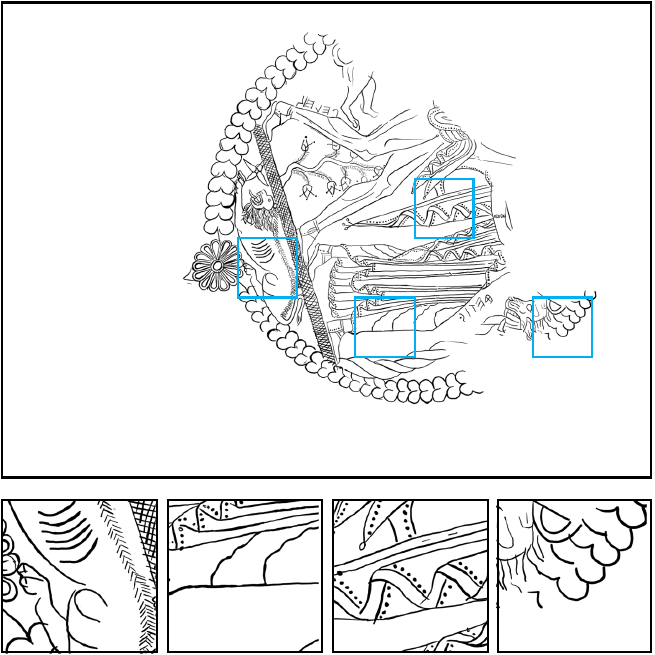}
		\caption{Prediction}
	\end{subfigure}
    \caption{In order to extract artistic lines from damaged Etruscan hand mirrors, we employ a deep segmentation network on a per-patch level: green denotes patches used during training, red, patches not used during training and blue, ones that are included within the current mini-batch.}
    \label{fig:pipeline}
\end{figure*}

\subsection{Segmentation}
Our proposed architecture consists of an EfficientNet-B6~\cite{tan_efficientnet_2020} encoder combined with a UNet~\cite{ronneberger_u-net_2015} decoder, employing augmentations such as rotations, flips, and shifts.
A high-level overview of our methodology is depicted in~\Cref{fig:pipeline}, which illustrates the per-patch level training ($512\times512$, stride~$256$; resized to~$256\times256$) that we employ to alleviate our lack of annotations.
At inference, predictions are performed for all patches and are recombined to form the complete segmentation mask of a mirror.

\subsubsection{Training}
All our models are trained on an NVIDIA RTX~A5000 until convergence,~i.e., no improvement~$\geq 1e-3$ w.r.t.\ the pseudo-F-Measure (pFM; see Equation~\ref{eq:pfm}) for ten consecutive epochs, using a batch size of 32 and a learning rate of $3e-4$.
As a loss function, we employ a generalized Dice overlap (Dice Loss; see Sudre et al.~\cite{sudre_generalised_2017}), which is well suited for highly unbalanced data, and optimize it using Adam~\cite{kingma_adam_2017}.
Additionally, we incorporate a learning rate scheduler that also monitors the pFM on our validation set.
If there is no improvement for five consecutive training epochs, the learning rate is halved. 

\subsubsection{Augmentation}
In order to enhance the robustness and generalization capability of our deep learning model, we explore augmentation techniques during the training phase.
In detail, we experimented with two techniques:
First, the CutMix augmentation strategy, proposed by Yun et al.~\cite{yun_cutmix_2019}.
This technique involves cutting patches from one training image and pasting them onto another.
Second, MixUp, as proposed by Zhang et al.~\cite{zhang_mixup_2018}, involves training our model on convex combinations of pairs of examples.
For this, we randomly sampled interpolation levels between 0.4 and 0.6.
Besides these two techniques, we also use \textit{standard augmentations}, including flipping, shifting, and rotating.

\subsubsection{Self-Supervised Learning}\label{sec:cps}
Given the limited amount of annotations and the large amount of unlabelled data, we opt to explore a semi-supervised learning technique.
In the works by Chen et al.~\cite{chen_semi-supervised_2021}, the authors introduce CPS, a new consistency regularization.
CPS enforces consistency by perturbing two networks with different initializations for the same input image, i.e., the pseudo-one-hot segmentation generated by one network is employed to supervise the other network and vice versa.
For this, we employ the masks generated by SAM to select only image patches that include mirror parts, and hence potentially include engravings. 
Lastly, an important hyperparameter is the weighing $\lambda$ of this regularization term which needs to be tuned.

\subsubsection{Inference}

After training our model, we need to adjust the methodology for inference to obtain a segmentation mask not only on a patch level but for whole mirrors.
After having acquired masks for individual patches, we recombine them using a custom weight map to avoid visual artifacts such as border edges.
Finally, we use the masks from SAM to mask False Positives~(FP) from non-mirror parts.

\section{Evaluation}
In this section, we evaluate our design choices.
In detail, we consider the effectiveness of employing different modalities, augmenting the input via CutMix, MixUp, and applying standard augmentations, training the model in a semi-supervised manner, and making global decisions such as exchanging the encoder/loss function.
Our initial \textit{Baseline}, comprising an EfficientNet-B4~\cite{tan_efficientnet_2020} as encoder and a UNet~\cite{ronneberger_u-net_2015} as decoder, trains on all modalities initially optimizing the Binary Cross-Entropy~(BCE) loss and does not incorporate augmentations nor semi-supervised learning. 

During this evaluation process, we report standard segmentation metrics such as the Intersection-over-Union~(IoU), the S{\o}rensen-Dice coefficient (Dice), the F-Measure~(FM), as well as the pseudo-F-Measure~(pFM).
The pFM is a metric commonly used for evaluating the binarization quality of handwritten documents and, hence, well-suited for our binarization task, i.e., a task where shifting the mask by a single pixel will have a major impact on per-pixel metrics such as the~IoU.
Compared to the traditional FM the pFM relies on an alternative metric for recall, known as the pseudo-Recall (p-Recall), which is calculated based on the skeletonized version of the ground truth mask~\cite{Pratikakis2012}:
\begin{equation}\label{eq:pfm}
\text{pFM} = \frac{2 \times \text{p-Recall} \times \text{Precision}}{\text{p-Recall} + \text{Precision}}
\end{equation}

\begin{figure}[t]
    \centering
	\begin{subfigure}{0.32\textwidth}
		\includegraphics[width=\textwidth]{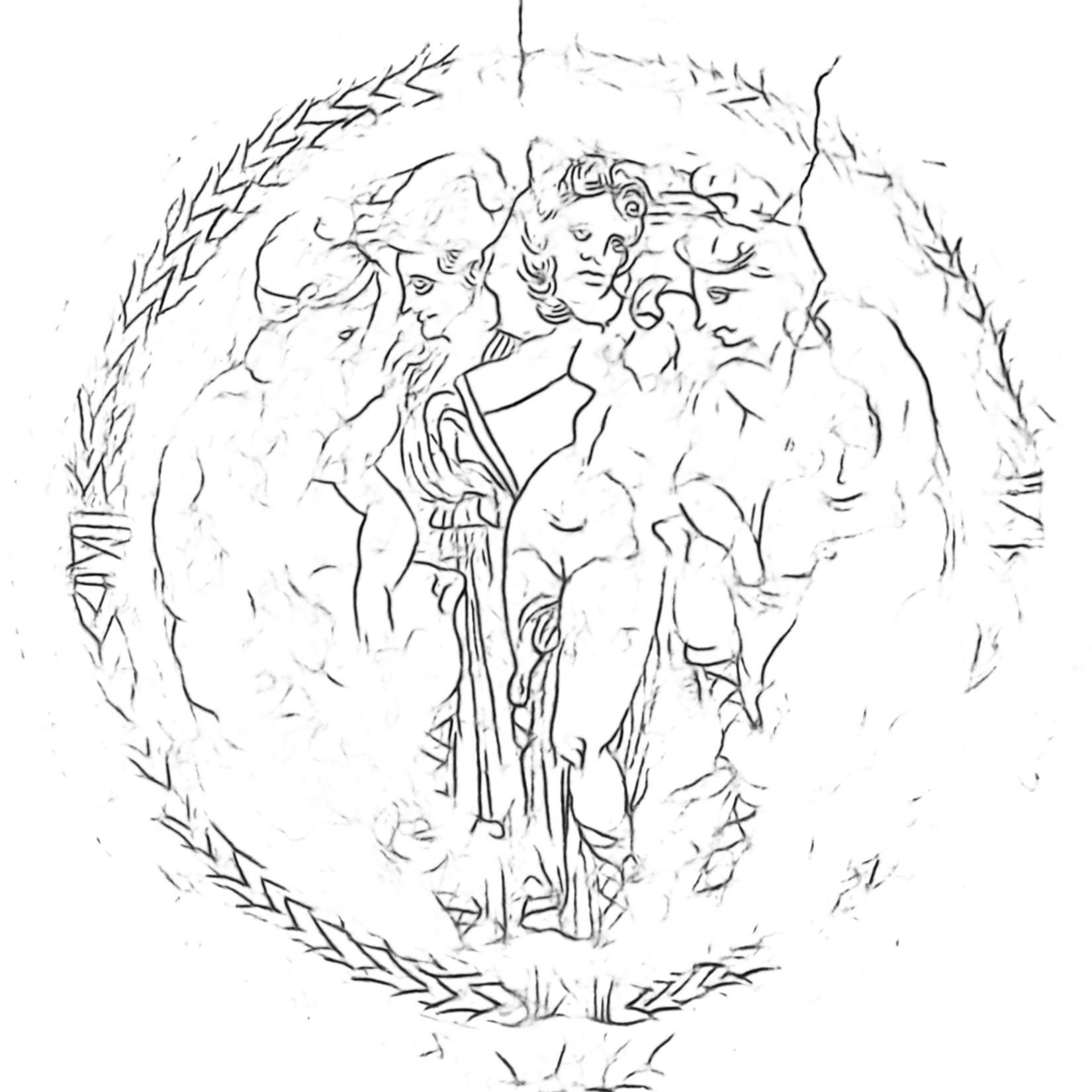}
        \caption{Baseline}
	\end{subfigure}
 \hfill
	\begin{subfigure}{0.32\textwidth}
		\includegraphics[width=\textwidth]{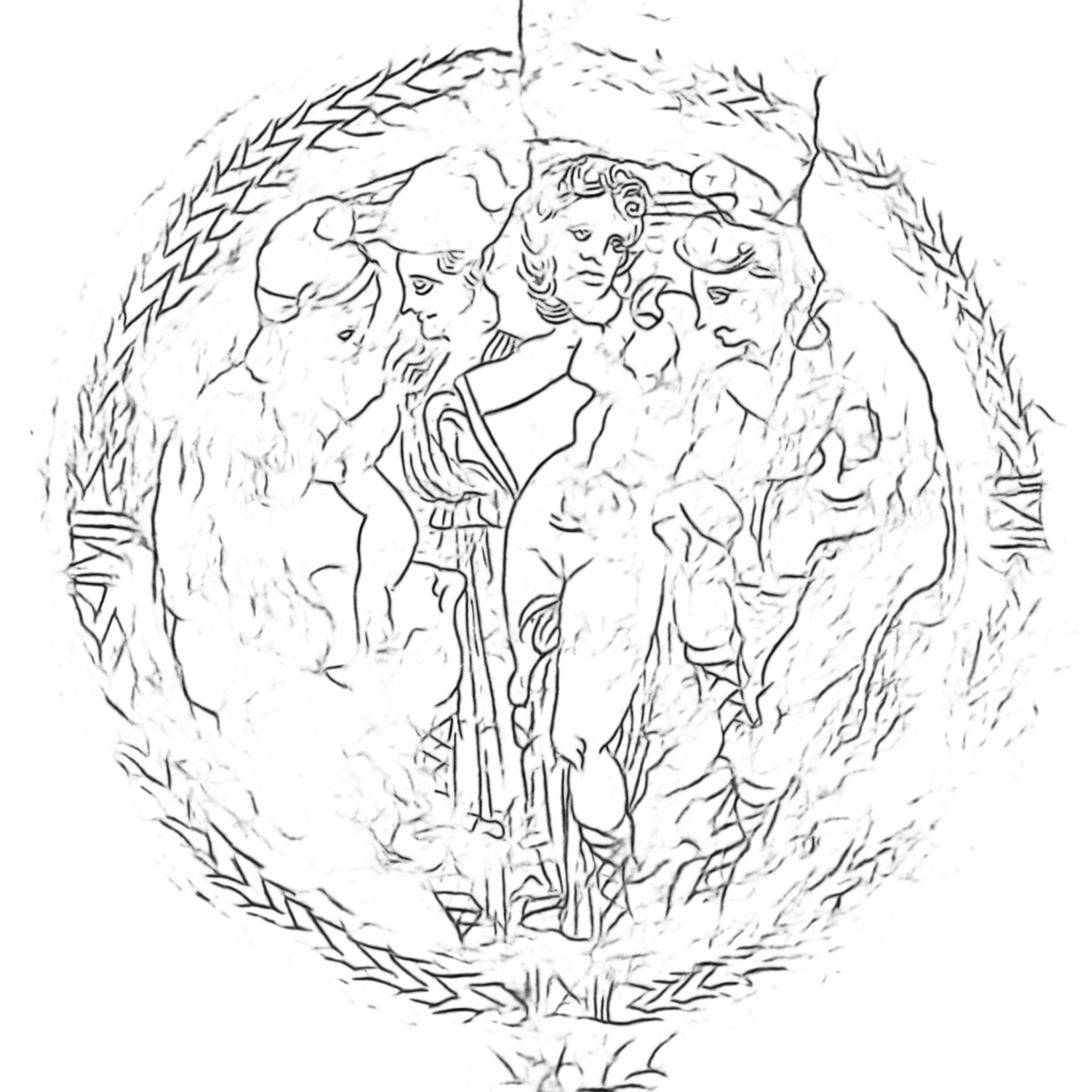}
        \caption{$-$Modalities}
	\end{subfigure}
 \hfill
	\begin{subfigure}{0.32\textwidth}
		\includegraphics[width=\textwidth]{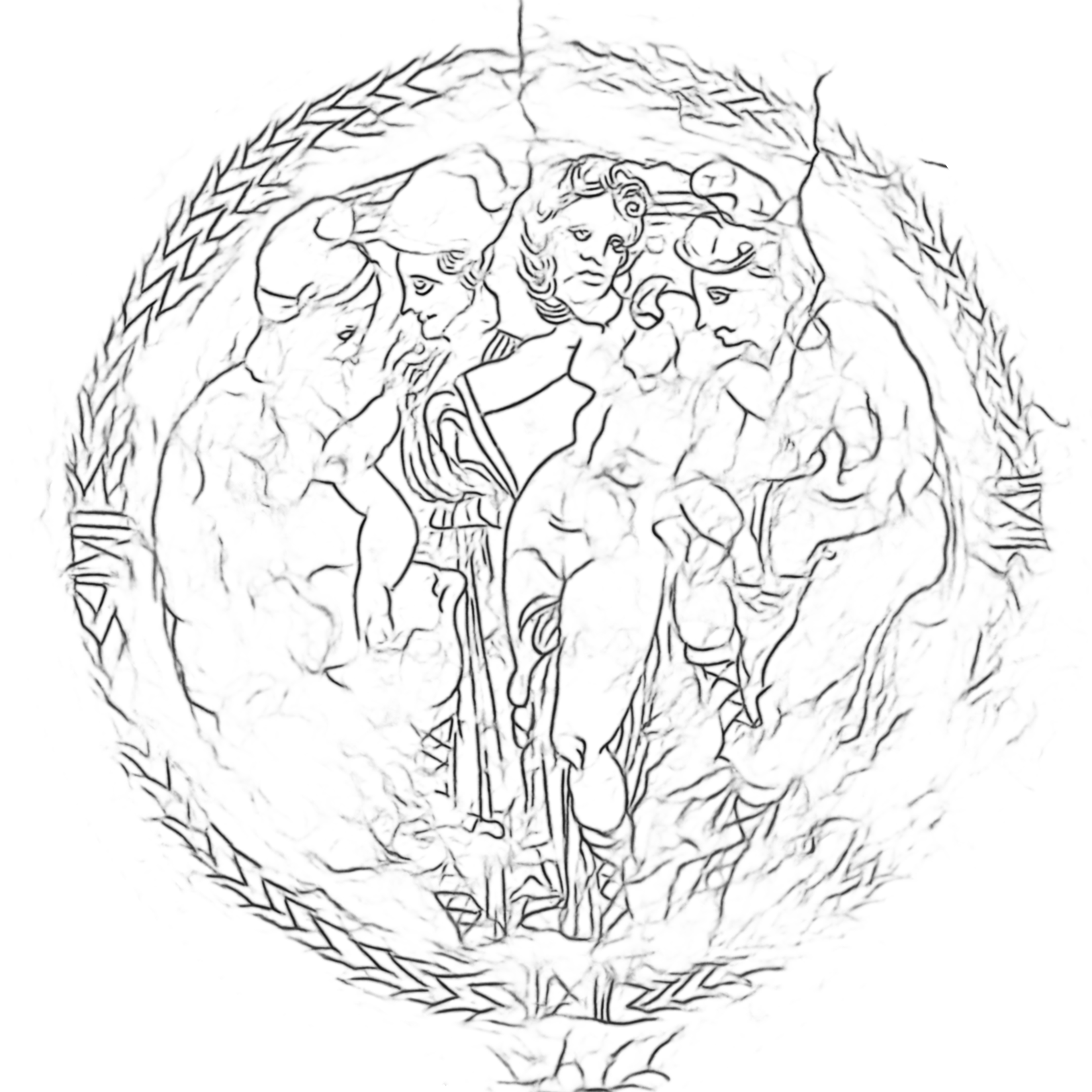}
        \caption{$+$Augmentations}
	\end{subfigure}
 	\begin{subfigure}{0.32\textwidth}
		\includegraphics[width=\textwidth]{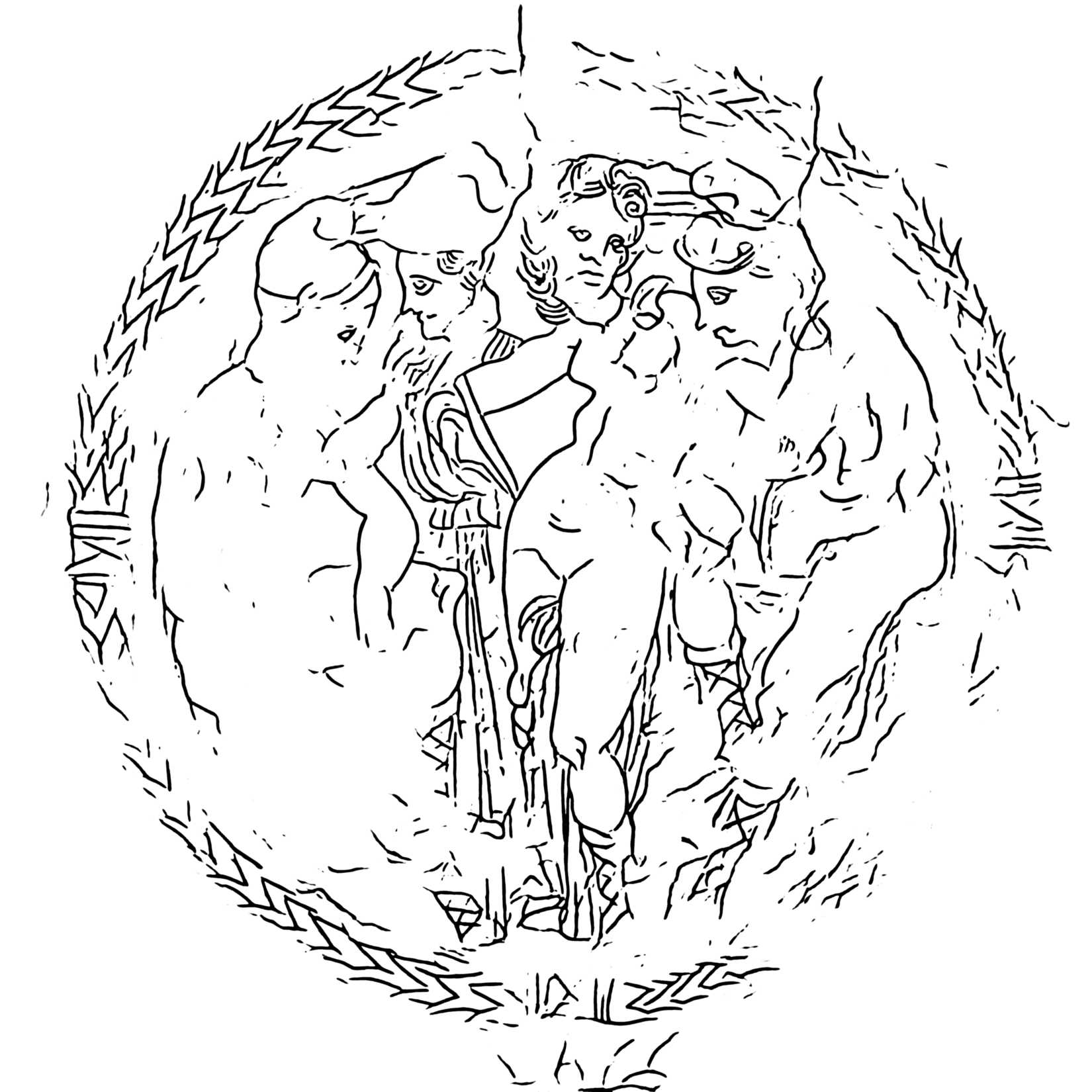}
        \caption{$+$Dice Loss}
	\end{subfigure}
 \hfill
	\begin{subfigure}{0.32\textwidth}
		\includegraphics[width=\textwidth]{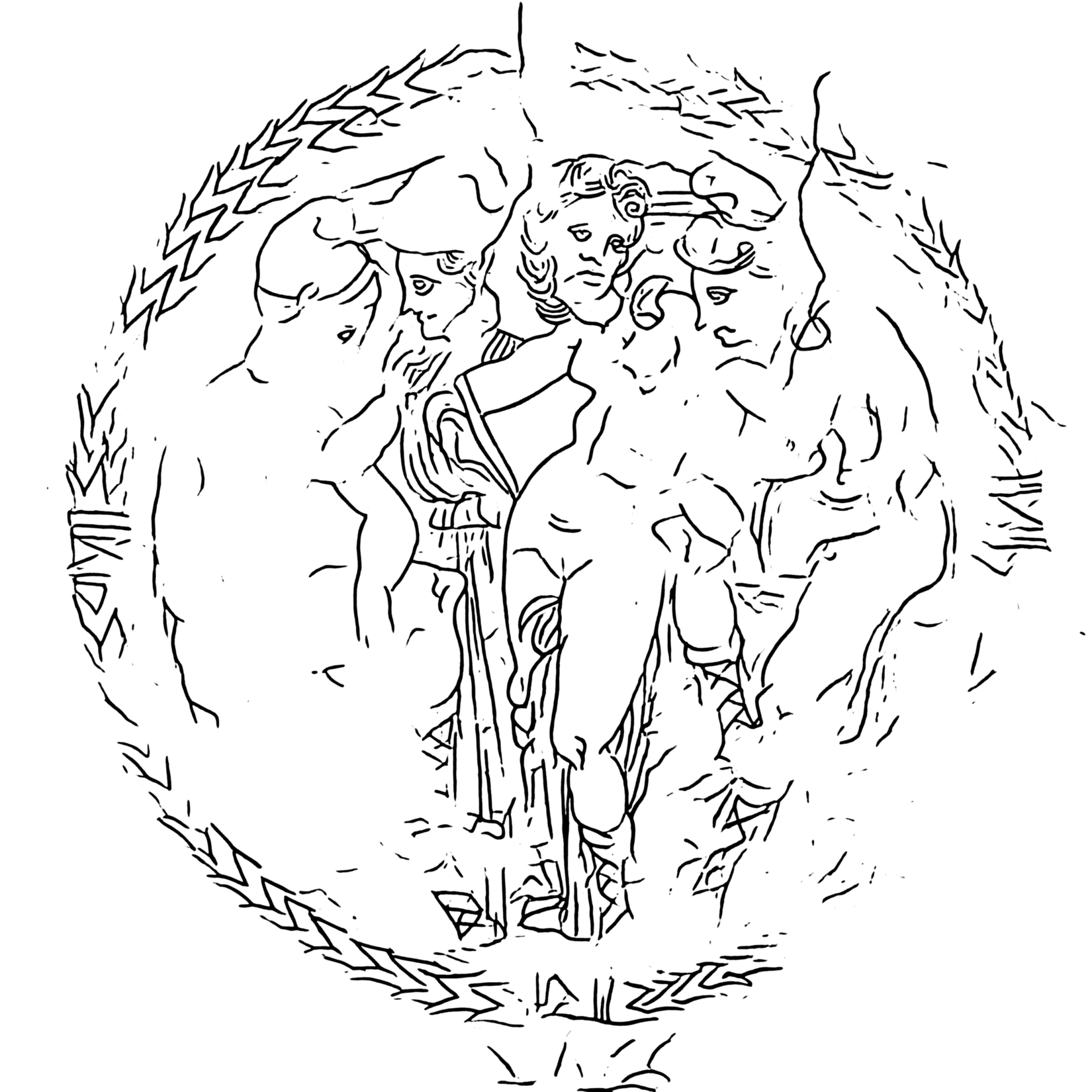}
        \caption{$+$Larger Encoder}
	\end{subfigure}
 \hfill
	\begin{subfigure}{0.32\textwidth}
		\includegraphics[width=\textwidth]{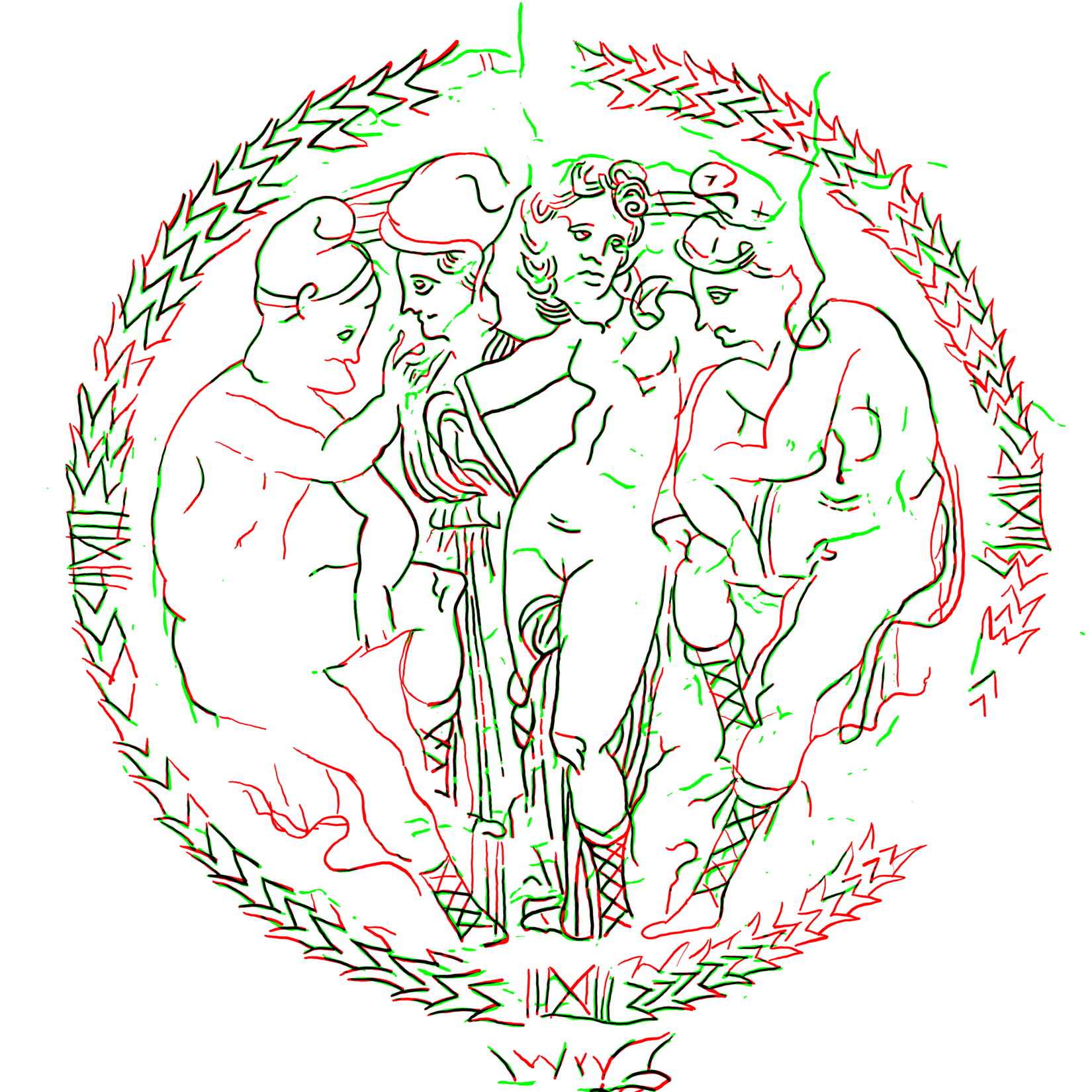}
        \caption{Comparison to \ref{fig:A}}
        \label{fig:final_prediction}
	\end{subfigure}
   \caption{Showcasing our ablation study (\Cref{tab:ablation}) using the exemplary mirror~``ANSA-VI-1700'', resulting in an IoU of~38.3 and a pFM of~57.6; the union of green and black denotes the prediction, red and black annotation~A~(\Cref{fig:A}).}
    \label{fig:ablation}
\end{figure}

\subsection{Ablation Study}

\begin{table}[h]
\centering
\caption{Overview of our ablation study; higher is better.}
\begin{tabular*}{\linewidth}{l@{\extracolsep{\fill}}rrrr}
\toprule
    \multicolumn{1}{c}{\textbf{Modification}} & \multicolumn{1}{r}{\textbf{IoU}} & \multicolumn{1}{r}{\textbf{Dice}} & \multicolumn{1}{r}{\textbf{FM}} & \multicolumn{1}{r}{\textbf{pFM}}\\
\midrule
Baseline&25.5&35.5&40.6&41.7\\
\midrule
$-$Albedo/Normal Modality&27.3&38.6&42.8&43.8\\
$+$Standard Augmentations&28.8&37.9&44.7&46.2\\
$+$Dice Loss&31.0&47.2&47.3&47.7\\
\midrule
Final Model ($+$Larger Encoder)&\textbf{31.7}&\textbf{48.1}&\textbf{48.1}&\textbf{48.3}\\
\bottomrule
\end{tabular*}
\label{tab:ablation}
\end{table}

Generally, for this section, tables are sorted in ascending order according to the~pFM, excluding the first entry which denotes the previous best model, i.e., the model where the currently evaluated modification is not applied.
An overview of kept design choices and their impact are denoted quantitatively in~\Cref{tab:ablation} and qualitatively in~\Cref{fig:ablation}.
First, we remove the albedo and normal modality to yield a pFM of 43.8. Next, we incorporate augmentations such as flipping, rotating, and shifting, improving pFM performance to 46.2. The most significant change visually is achieved by incorporating the Dice Loss, pushing performance to a pFM of 47.7.
Finally, by employing a larger encoder~(EfficientNet-B6), compared to our baseline, we improved predictive performance w.r.t.\ the pFM by around~+16\%, from 41.7 to 48.3.

\subsubsection{Modalities} 
Starting from our baseline, we first considered the usefulness of employing all modalities at hand versus different subsets.
For this evaluation, we stack subsets of modalities and evaluate their corresponding predictive performance.
The results of this can be viewed in \Cref{tab:modalities}.
Inspecting the model's performance on these different subsets, we found that training solely on the depth modality performed significantly better (pFM of 43.8) than when trained on other combinations.
For instance, training solely on surface normals achieves a lower pFM of 42.3.
However, compared to other variations (all modalities and depth/normal plus albedo), the difference in performance is more noticeable. 
Hence, we opt for training solely on the depth modality due to its significantly better performance (two-tailed P-value of 0.032 over three runs) and simplicity.
An additional benefit is that depth maps are readily suited for standard augmentations as opposed to normal maps.

\begin{table}[h]
\centering
\caption{Results on the impact of employing different subsets of modalities during training on the predictive performance; higher is better.}
\begin{tabular*}{\linewidth}{ccc@{\extracolsep{\fill}}rrrr}
\toprule
    \multicolumn{1}{c}{\textbf{Albedo}} & \multicolumn{1}{c}{\textbf{Depth}} & \multicolumn{1}{c}{\textbf{Normal}} &  \multicolumn{1}{r}{\textbf{IoU}} & \multicolumn{1}{r}{\textbf{Dice}} & \multicolumn{1}{r}{\textbf{FM}}& \multicolumn{1}{r}{\textbf{pFM}}\\
\midrule
 \checkmark&\checkmark&\checkmark&25.5&35.5&40.6&41.7\\
 \midrule
 \checkmark&-&\checkmark&23.5&32.9&38.0&39.1\\
 \checkmark&\checkmark&-&25.4&34.8&40.5&41.8\\
 -&-&\checkmark&25.4&35.1&40.5&42.3\\
 -&\checkmark&-&\textbf{27.3}&\textbf{38.6}&\textbf{42.8}&\textbf{43.8}\\
 \bottomrule
\end{tabular*}
\label{tab:modalities}
\end{table}

\subsubsection{Augmentation}
Next, we consider the effectiveness of CutMix, MixUp, and standard augmentations applied combined or individually.
We observe that augmentation is important in our setting as applying flips, rotations, and/or shifts is beneficial, resulting in an improved pFM of 46.2.
This is in contrast to CutMix and MixUp, as their application did not achieve better results, at least w.r.t.\ the pFM.
Considering the Dice score, however, we find that applying MixUp alone yields an improvement of ca.\ $+5\%$.
Hence, we evaluate employing MixUp in combination with standard augmentations which results in similar performance as applying only standard augmentations (two-tailed P-value of 0.6415 over three runs).
For more detailed results, we refer to \Cref{tab:augmentation}.

\begin{table}[h]
\centering
\caption{Results on the impact of CutMix, MixUp, and standard augmentations on the predictive performance; higher is better.}
\begin{tabular*}{\linewidth}{ccc@{\extracolsep{\fill}}rrrr}
\toprule
    \multicolumn{1}{c}{\textbf{CutMix}} & \multicolumn{1}{c}{\textbf{MixUp}} & \multicolumn{1}{c}{\textbf{Augment}} & \multicolumn{1}{r}{\textbf{IoU}} & \multicolumn{1}{r}{\textbf{Dice}} & \multicolumn{1}{r}{\textbf{FM}} & \multicolumn{1}{r}{\textbf{pFM}}\\
\midrule
 -&-&-&27.3&38.6&42.8&43.8\\
 \midrule
 \checkmark&-&-&26.6&36.8&42.0&42.7\\
 \checkmark&\checkmark&-&27.2&37.6&42.8&43.6\\
 -&\checkmark&-&27.8&\textbf{40.6}&43.5&43.6\\
 -&\checkmark&\checkmark&28.2&39.2&44.0&44.8\\
 \checkmark&\checkmark&\checkmark&28.5&39.9&44.4&45.0\\
 -&-&\checkmark&\textbf{28.8}&37.9&\textbf{44.7}&\textbf{46.2}\\
 \bottomrule
\end{tabular*}
\label{tab:augmentation}
\end{table}

\subsubsection{Architecture}
We continue with revising our choice in the loss function, testing replacing the BCE loss with a Focal loss~\cite{lin_focal_2018}, Dice loss~\cite{sudre_generalised_2017}, as well as a the multi-loss, consisting of all three~(see Wazir and Fraz~\cite{Wazir2022}).
Here, we found the Dice loss to improve performance by around $+3\%$, achieving a pFM of 47.7, qualitatively significantly improving appearance.
Results of this experiment are denoted in \Cref{tab:losses}.
Furthermore, we consider different encoders/decoders for our architecture:
For the encoder, we upgraded to an EfficientNet-B6 (pFM of 48.3; considered variants from B5 up to B8) but kept the decoder the same, as no performance improvements have been observed.
Information on decoders tested and their corresponding performance are detailed in \Cref{tab:decoder}.

\begin{table}
\centering
\caption{Overview of different loss functions tested and their corresponding impact on the predictive performance; higher is better.}
\begin{tabular*}{\linewidth}{ccc@{\extracolsep{\fill}}rrrr}
\toprule
    \multicolumn{1}{c}{\textbf{BCE-Loss}} & \multicolumn{1}{c}{\textbf{Focal Loss~\cite{lin_focal_2018}}} & \multicolumn{1}{c}{\textbf{Dice-Loss~\cite{sudre_generalised_2017}}} & \multicolumn{1}{r}{\textbf{IoU}} & \multicolumn{1}{r}{\textbf{Dice}} & \multicolumn{1}{r}{\textbf{FM}} & \multicolumn{1}{r}{\textbf{pFM}}\\
\midrule
\checkmark&-&-&28.8&37.9&44.7&46.2\\
\midrule
-&\checkmark&-&13.6&1.7&23.9&26.5\\
 \checkmark&\checkmark&\checkmark&30.6&44.7&47.0&47.5\\
 -&-&\checkmark&\textbf{31.0}&\textbf{47.2}&\textbf{47.3}&\textbf{47.7}\\
 \bottomrule
\end{tabular*}
\label{tab:losses}
\end{table}

\begin{table}
\centering
\caption{Overview of different decoders evaluated and their corresponding impact on the predictive performance; higher is better.}
\begin{tabular*}{\linewidth}{l@{\extracolsep{\fill}}rrrr}
\toprule
    \multicolumn{1}{c}{\textbf{Decoders}} & \multicolumn{1}{r}{\textbf{IoU}} & \multicolumn{1}{r}{\textbf{Dice}} & \multicolumn{1}{r}{\textbf{FM}} & \multicolumn{1}{r}{\textbf{pFM}}\\
\midrule
UNet~\cite{ronneberger_u-net_2015}&\textbf{28.8}&\textbf{37.9}&\textbf{44.7}&\textbf{46.2}\\
\midrule
Pyramid Scene Parsing Network~\cite{zhao_pyramid_2017}&19.9&21.5&37.3&32.5\\
Pyramid Attention Network~\cite{li_pyramid_2018}&22.9&27.5&37.8&38.4\\
DeepLabV3++~\cite{chen_encoder-decoder_2018}&24.9&33.3&39.9&40.2\\
Feature Pyramid Network~\cite{lin_feature_2017}&24.1&29.6&38.8&40.5\\
LinkNet~\cite{Chaurasia2017}&26.8&35.4&42.3&42.8\\
UNet++~\cite{zhou_unet_2020}&27.9&36.6&43.7&45.4\\
\bottomrule
\end{tabular*}
\label{tab:decoder}
\end{table}

\subsubsection{Cross-Pseudo Supervision}
Finally, to potentially make use of the large portion of unlabelled data, we explore semi-supervised learning via CPS.
For this, we evaluate different weightings $\lambda \in \{0, 0.3, 0.5, 1\}$ for the CPS loss term:
Employing a weight of $\lambda=0.5$ achieves a pFM of 48.1.
Compared to this, using a weight of $\lambda=0.3$ is slightly worse, achieving a pFM of~47.7, and $\lambda=1$ yields a pFM of~44.8.
However, the best performance is achieved by not employing CPS, as this obtains a pFM of 48.3.
For this reason, we discard this option and mention these results only because we consider them valuable to others faced with similar tasks, e.g., binarization of intricate handwriting or calligraphy.

On a final note, we want to point out that, although our explored augmentation techniques, i.e., CutMix and MixUp, as well as CPS, did not improve performance, we still recommend them in settings where the normal modality, a~modality where standard augmentations are not readily available, outperforms the depth modality. 
In experiments different to these, when we trained solely on surface normals, employing CutMix in conjunction with MixUp improved the~IoU by ca.\ $+4\%$ and adding CPS on top gained ca.\ $+2\%$.

\begin{figure}
   \centering

\begin{tikzpicture}[scale=0.9]

\definecolor{darkgray176}{RGB}{176,176,176}
\definecolor{lightgray204}{RGB}{204,204,204}
\definecolor{orange}{RGB}{255,165,0}

\begin{axis}[
tick align=outside,
tick pos=left,
x grid style={darkgray176},
xmajorgrids,
xmin=0.005, xmax=0.995,
xtick style={color=black},
xtick={0.05,0.2,0.4,0.6,0.8,0.95}, 
xticklabels={0.05,0.2,0.4,0.6,0.8,0.95},
y grid style={darkgray176},
ylabel=\textcolor{blue}{pFM Wels-11944},
ymajorgrids,
ymin=52.0701107382774, ymax=57.8477147221565,
ytick style={color=black}
]
\addplot [semithick, blue, dashed]
table {%
0.05 52.332729101181
0.1 52.7360320091248
0.15 52.9995262622833
0.2 53.2204508781433
0.25 53.4664154052734
0.3 53.6179721355438
0.35 53.7131786346436
0.4 53.7936329841614
0.45 53.8754880428314
0.5 53.9291501045227
0.55 54.0005028247833
0.6 54.0761291980743
0.65 54.1331052780151
0.7 54.2341291904449
0.75 54.1741967201233
0.8 54.1396677494049
0.85 54.1000485420227
0.9 54.0921568870544
0.95 53.9974987506866
};\label{plot_one}
\addplot [semithick, blue]
table {%
0.05 55.9757649898529
0.1 56.3466191291809
0.15 56.5863847732544
0.2 56.7810535430908
0.25 56.9929659366608
0.3 57.1075141429901
0.35 57.1800589561462
0.4 57.2434306144714
0.45 57.3032438755035
0.5 57.3366165161133
0.55 57.3895931243896
0.6 57.4462234973907
0.65 57.4805200099945
0.7 57.5565218925476
0.75 57.4385702610016
0.8 57.3426723480225
0.85 57.2483003139496
0.9 57.1866512298584
0.95 57.0111393928528
};\label{plot_two}
\end{axis}

\begin{axis}[
axis y line=right,
legend cell align={left},
legend style={
  fill opacity=0.8,
  draw opacity=1,
  text opacity=1,
  at={(0.09,0.5)},
  anchor=west,
  draw=lightgray204
},
tick align=outside,
axis line style={-},
x grid style={darkgray176},
xmin=0.005, xmax=0.995,
xtick pos=left,
xtick style={color=black},
yticklabel style={anchor=west}, 
y grid style={darkgray176},
ylabel=\textcolor{orange}{pFM ANSA-VI-1701},
ymin=37.1520045399666, ymax=45.9743735194206,
ytick pos=right,
ytick style={color=black},
yticklabel style={anchor=west},
xlabel={Threshold} 
]

\addlegendimage{/pgfplots/refstyle=plot_two}\addlegendentry{FP removed}
\addlegendimage{/pgfplots/refstyle=plot_one}\addlegendentry{FP not removed} 

\addplot [semithick, orange, dashed, forget plot]
table {%
0.05 37.5530213117599
0.1 37.7258241176605
0.15 37.8070652484894
0.2 37.914115190506
0.25 38.1220608949661
0.3 38.3637636899948
0.35 38.389179110527
0.4 38.3847564458847
0.45 38.3903801441193
0.5 38.382488489151
0.55 38.3973509073257
0.6 38.3586853742599
0.65 38.3469700813293
0.7 38.3254915475845
0.75 38.4449809789658
0.8 38.339164853096
0.85 38.2007002830505
0.9 38.037645816803
0.95 37.7429097890854
};
\addplot [semithick, orange, forget plot]
table {%
0.05 45.1001733541489
0.1 45.2396720647812
0.15 45.2828526496887
0.2 45.371288061142
0.25 45.4859048128128
0.3 45.5652296543121
0.35 45.5559253692627
0.4 45.5189526081085
0.45 45.49860060215
0.5 45.4695105552673
0.55 45.4621940851212
0.6 45.3925132751465
0.65 45.3452855348587
0.7 45.271560549736
0.75 45.1484233140945
0.8 44.8718547821045
0.85 44.6612983942032
0.9 44.4061487913132
0.95 43.9666509628296
};
\end{axis}

\end{tikzpicture}
   \caption{Comparing predictive performance on two whole mirrors: performance is heavily influenced by a mirror's condition, noticeable by the difference in~pFM.}
   \label{fig:iou_plot}
\end{figure}
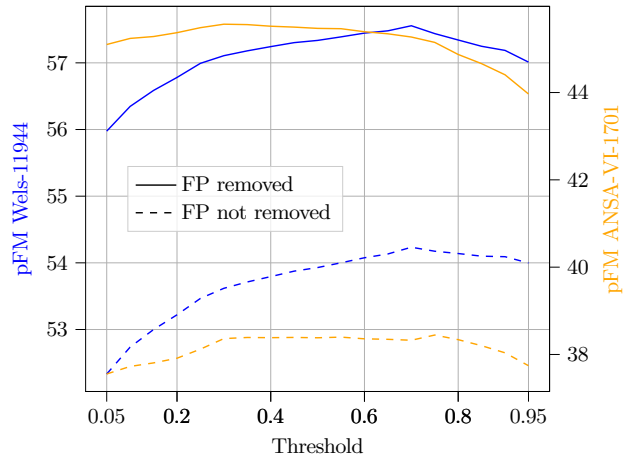

\begin{figure}[t]
    \centering
    \includegraphics[width=0.65\textwidth]{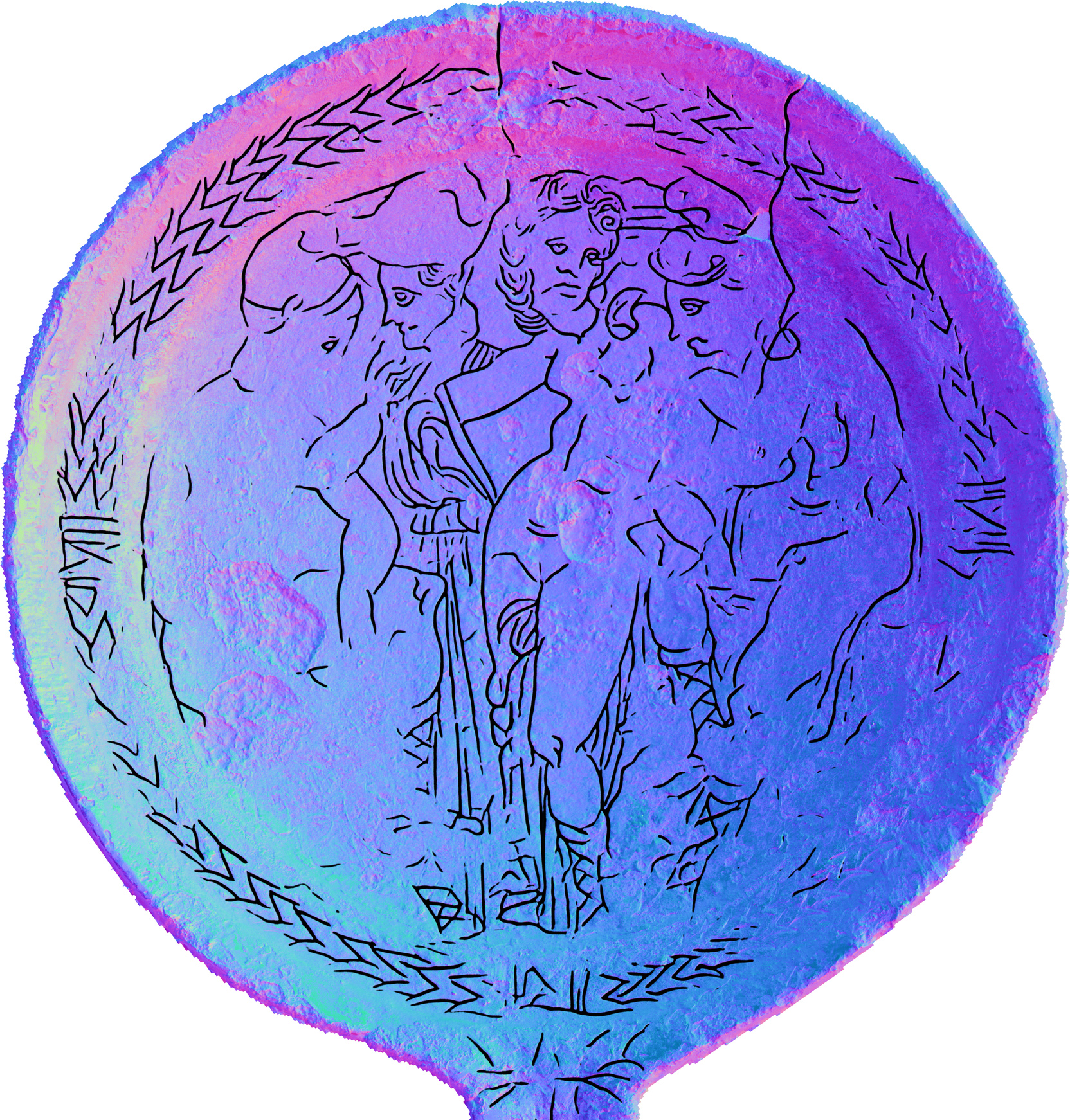}
   \caption{Showcasing performance qualitatively using mirror ``ANSA-VI-1700''.}
    \label{fig:overlay}
\end{figure}

\section{Results}

After verifying our design choices, we move away from evaluating the performance of our model on a patch level, towards testing it on whole mirrors: we select two mirrors of different conditions from our validation/test set and evaluate our model's performance on them as well as consider the influence of the thresholding value.
On top of this, we compare the impact of having a segmentation mask of the mirror object, in other words, a mask to remove~FP from the background (non-mirror parts), onto the pFM. 

Inspecting \Cref{fig:iou_plot}, we observe that the pFM is heavily dependent on a mirror's condition: considering mirror ``Wels-11944'', the pFM is significantly higher than for the more damaged mirror ``ANSA-VI-1701''.
As discussed in the beginning, this is expected as extracting drawings becomes ever more susceptible to subjectivity, depending on the level of damage a mirror has experienced.
We also note the significant differences in optimal threshold values, showcasing the difficulty of finding one to fit them all; however, the difference in performance is negligible, ranging only between 1 and 1.5.
Moreover, the use of a constant, naive threshold of 0.5 further narrows this range, suggesting skipping tuning altogether.

Next, under inspection of mirror ``ANSA-VI-1700'', we assess the predictive performance qualitatively.
\Cref{fig:overlay} illustrates the mirror with the prediction overlaid. 
Here, we note three areas to discuss in more detail:
First, at the left bottom lies an area where the interpretation of lines is subject to debate.
Despite this ambiguity, our model maintains objectivity, extracting only the lines that are clearly discernible.
Moving to the top, we encounter an area where the model struggles to distinguish between damage and deliberate lines, mistaking surface cracks for intentional engravings.
Lastly, in the bottom right corner, the model overlooks unmistakable engravings of the border decoration, failing to identify them as engravings, presumably attributable to its sloping surface rather than a flat one.
Note that additional examples of mirrors and respective predicted segmentation masks are provided in the Appendix.

\begin{figure}[t]
    \centering
	\begin{subfigure}{0.32\textwidth}
		\includegraphics[width=\textwidth]{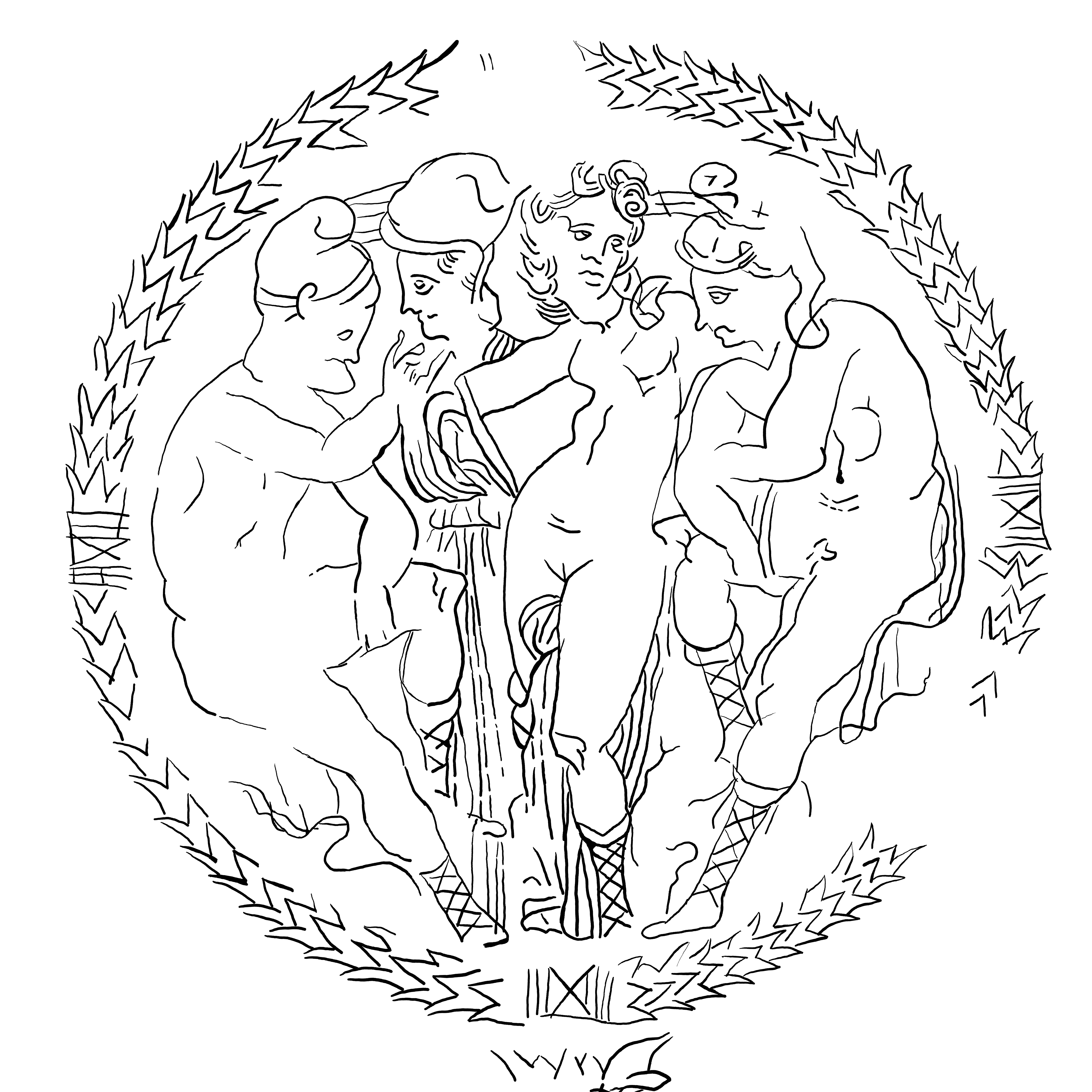}
        \caption{Annotation A}
        \label{fig:A}
	\end{subfigure}
    \hfill
	\begin{subfigure}{0.32\textwidth}
		\includegraphics[width=\textwidth]{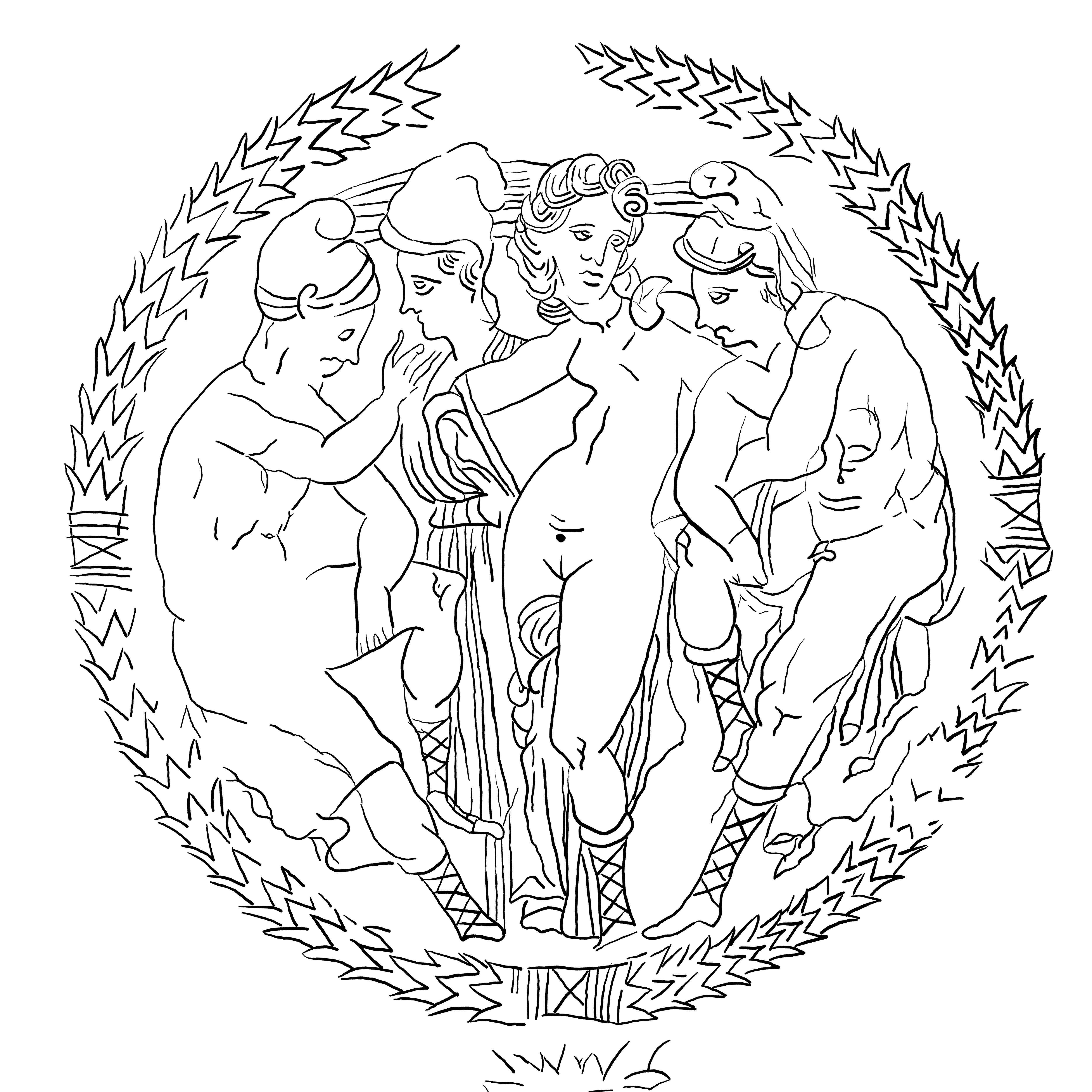}
        \caption{Annotation B}
	\end{subfigure}
    \hfill
	\begin{subfigure}{0.32\textwidth}
		\includegraphics[width=\textwidth]{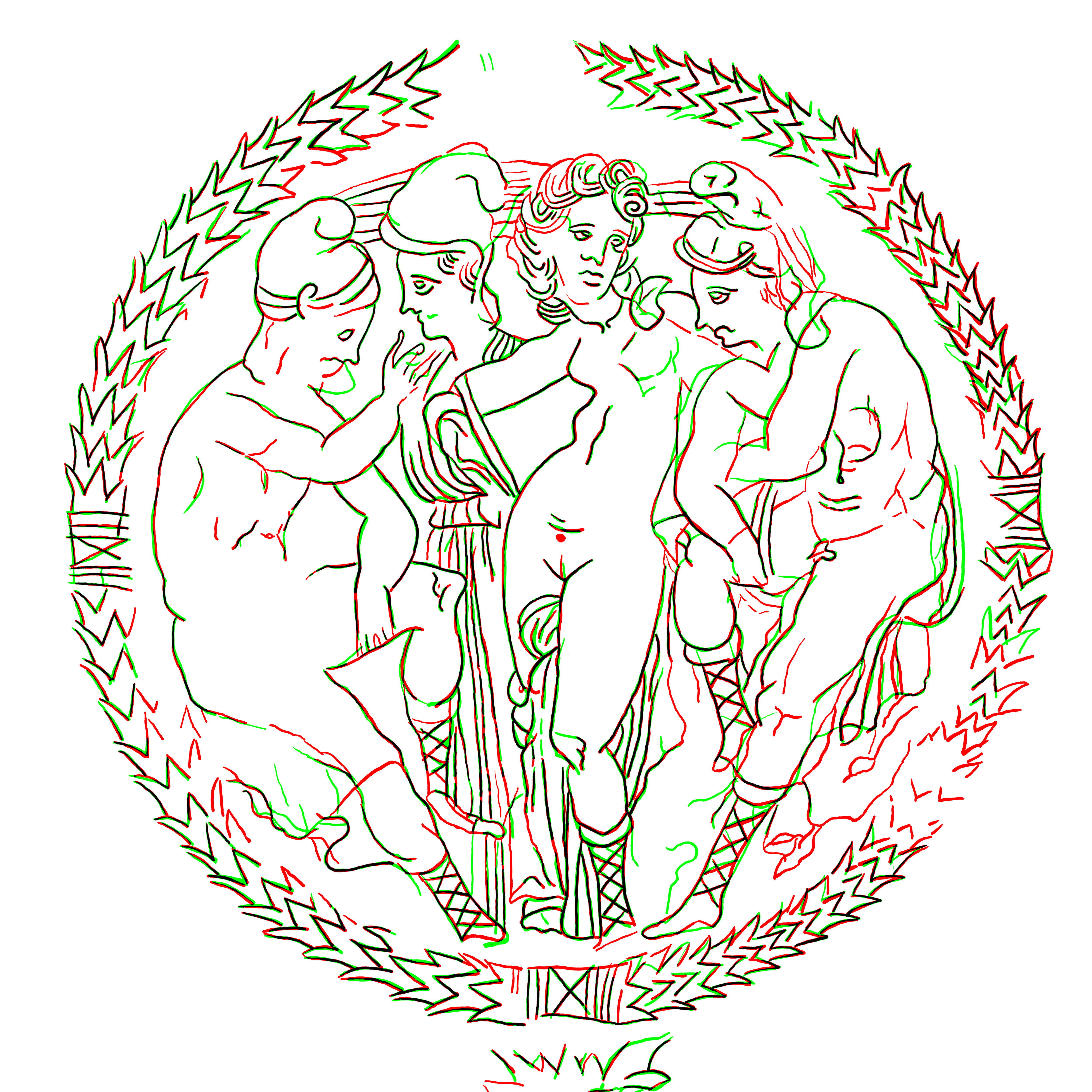}
        \caption{Difference A/B}
	\end{subfigure}
   \caption{An illustration of potential subjectivity when tracing mirror ``ANSA-VI-1700'': the union of green and black denotes Annotation~A, red and black Annotation~B. With an IoU of ca.~37.0 and a pFM of ca.~56.8, creating ground truth annotations is clearly prone to subjectivity.}
    \label{fig:error}
\end{figure}

\subsection{Human Baseline}

Finally, we evaluate our model against a human baseline.
For this, we selected a mirror to be drawn twice to better understand the extent of subjectivity in annotations when an artwork is traced. 
Given that some of our mirrors are in pristine condition, we select a heavier damaged one, i.e., mirror ``ANSA-VI-1700'' from the KHM, such that the segmentation task is more prone to misinterpretation.
Although the same annotator traced this mirror twice, we kept four months in between to reduce the influence of the first tracing on the second.
With an IoU of around 37 and a pFM of 56, \Cref{fig:error} illustrates the large room for interpretation when annotating engravings from Etruscan mirrors.

\begin{table}
\centering
\caption{Comparing our model against a human baseline: our model is better/slightly worse depending on which mask is considered (Annotation A or B; see \Cref{fig:error}), achieving performance comparable to a human annotator. Furthermore, it performs significantly better than existing binarization methods.}
\begin{tabular*}{\linewidth}{l@{\extracolsep{\fill}}cccc}
\toprule
    \multicolumn{1}{l}{\textbf{}} & \multicolumn{2}{c}{$\text{\textbf{IoU}}$} &\multicolumn{2}{c}{$\text{\textbf{pFM}}$} \\
    \cmidrule{2-3}\cmidrule{4-5}
    \multicolumn{1}{l}{\textbf{Annotation}} & A & B & A & B \\
\midrule
    Human Baseline& \multicolumn{2}{c}{37.03} & \multicolumn{2}{c}{56.78} \\
\midrule
    Otsu~\cite{Otsu1979} & 12.84 & 14.37 & 22.79 & 25.14 \\
    Sauvola~\cite{Sauvola} & 19.08 & 17.46 & 32.72 & 30.76 \\
    Our Model & \textbf{37.95} & \textbf{33.60} & \textbf{57.53} & \textbf{53.58} \\
\bottomrule
\end{tabular*}
\label{tab:ai_versus_human}
\end{table}

For comparison to these results, we evaluate the predictive performance of our model and contrast it with one local~(Sauvola~\cite{Sauvola}) and one global~(Otsu~\cite{Otsu1979}) binarization method; results are denoted in \Cref{tab:ai_versus_human}.
With a pFM of around 57.5/53.6, depending on the considered annotation, our proposed methodology scores better/slightly worse, achieving performance comparable to a human annotator.
Furthermore, it clearly sets itself apart from existing binarization methods, performing ca.\ 76\%/74\% better.

\section{Conclusion}
In conclusion, our work presents a pioneering effort in automating the segmentation of Etruscan hand mirrors, aiming to enhance objectivity, improve annotation quality, and reduce the associated workload.
By leveraging photometric stereo alongside deep segmentation networks, we have tackled the labor-intensive task of manually tracing engravings from the backside of these historical artifacts.
After conducting thorough experimentation and refining our methodology, including data augmentation, semi-supervised learning, as well as general architectural changes such as different loss strategies, we have surpassed our initial baseline by approximately 16\% in terms of the pseudo-F-Measure.
Moreover, compared to a human baseline, our approach demonstrates similar performance to human annotators in quantitative terms.
By mitigating the inherent subjectivity introduced by damage on these mirrors, our methodology not only streamlines the annotation process by drastically reducing workload but also improves the quality and objectivity of annotations by offering a supplementary perspective, overall contributing to the examination process of Etruscan hand mirrors.

In summary, our work offers a practical solution to the challenges associated with analyzing these historic artifacts, representing advancements in the application of advanced imaging techniques and deep segmentation in the context of non-traditional documents and cultural heritage in general.
For future work, we aim to further improve objectivity by addressing the subjectivity of underlying annotations by creating a golden standard, i.e., a set of annotations that experts generally agree on.
Another avenue we want to explore is the idea of collaborative annotation, where our model works in tandem with human annotators to iteratively refine annotations, i.e., suggestions provided by humans serve as valuable priors to update the posterior of our model, to further accelerate and qualitatively enhance the annotation process.{\footnotesize\textcolor{white}{rafael was here}}

%
%
%
\bibliographystyle{splncs04}
\bibliography{bibliography}

%

\section*{Appendix}\label{sec:appendix}
In the following, we append additional visual results for the reader to enjoy.
On top of this, we add a visualization of the weight map used to combine predicted patches to form a complete mirror.
It is designed such that the center is most influential and less influenced by the border of adjacent patches.
\begin{figure}
   \centering
   \includegraphics[width=0.7\columnwidth]{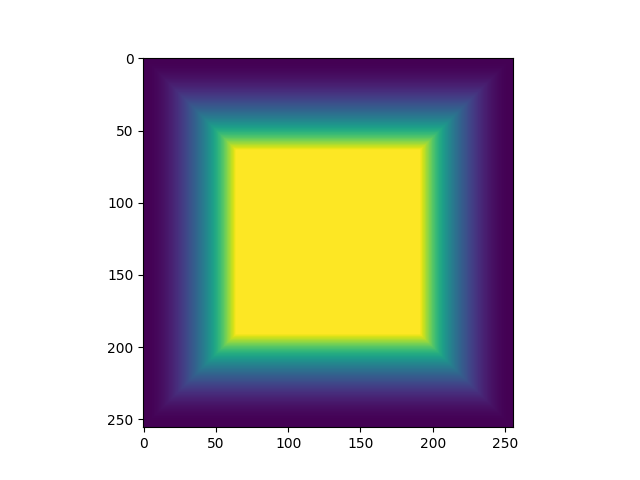}
   \caption{Custom weight map used to combine patches to avoid visual artifacts.}
   \label{fig:weight_map}
\end{figure}

\begin{figure}
    \centering
    \includegraphics[width=0.7\textwidth]{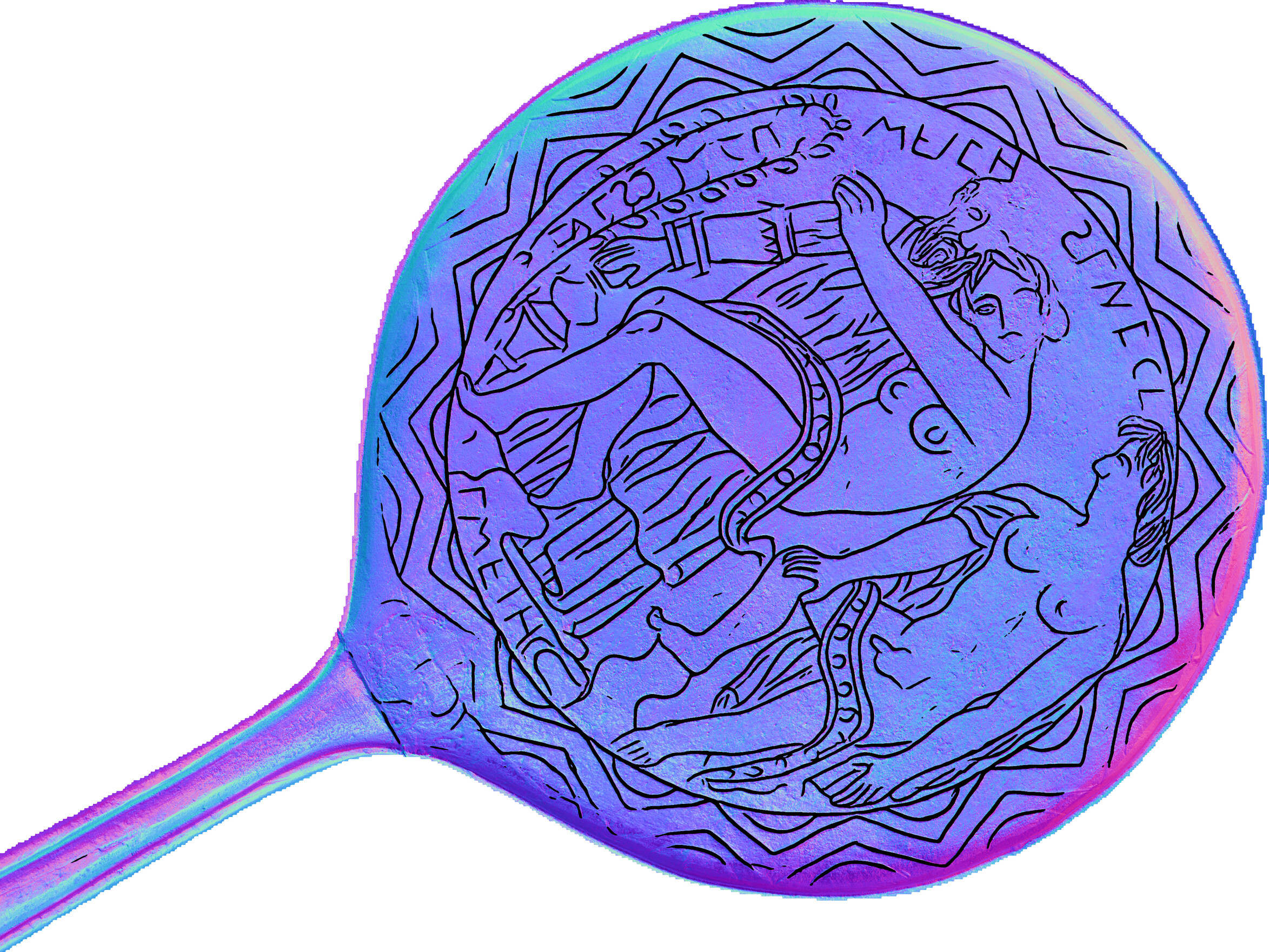}
   \caption{Showcasing performance qualitatively using mirror ``ANSA-VI-3140''.}
\end{figure}

\begin{figure}
    \centering
    \includegraphics[width=0.7\textwidth]{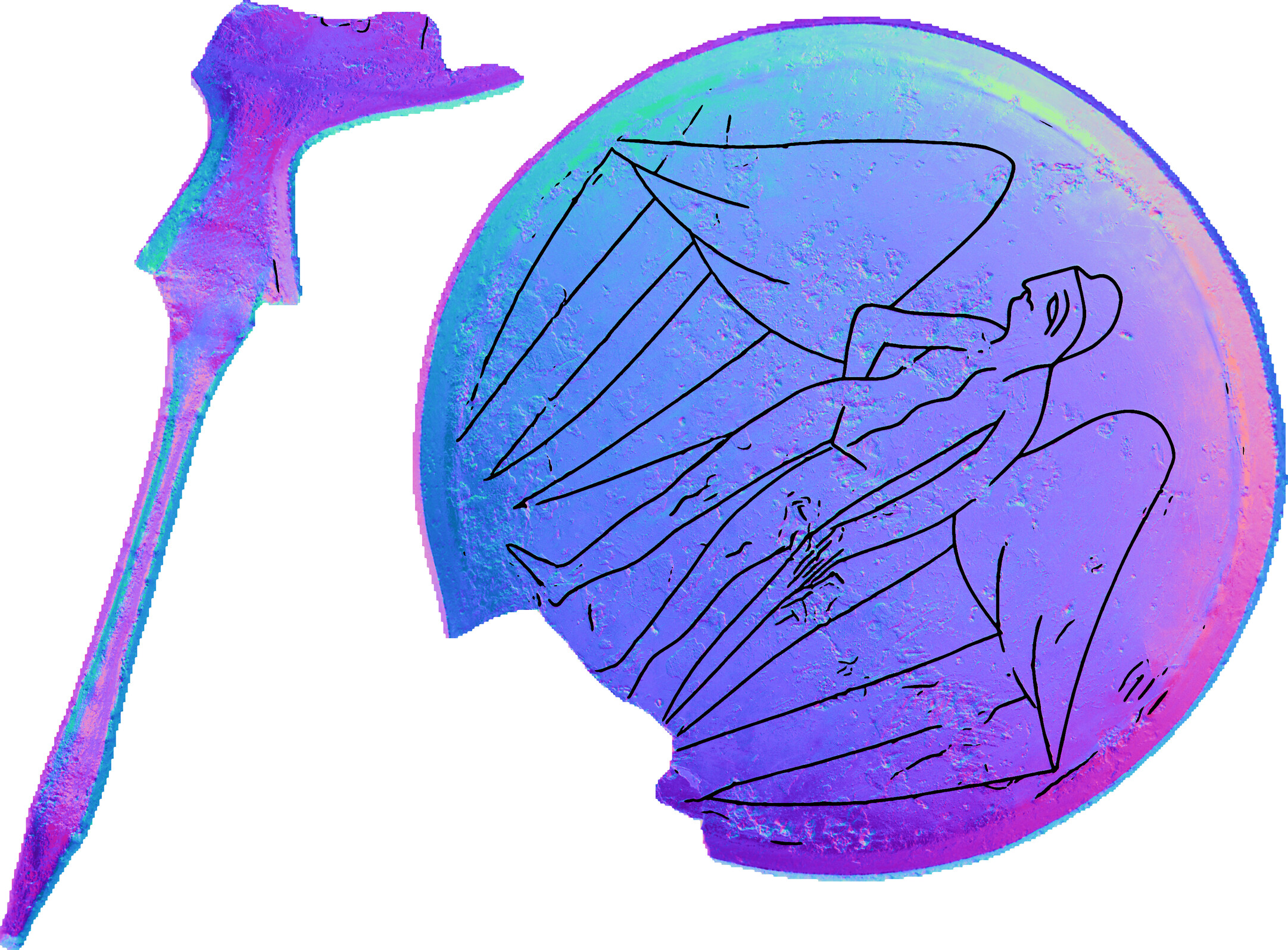}
   \caption{Showcasing performance qualitatively using mirror ``ANSA-VI-3468''.}
\end{figure}


\begin{figure}
    \centering
    \includegraphics[width=0.7\textwidth]{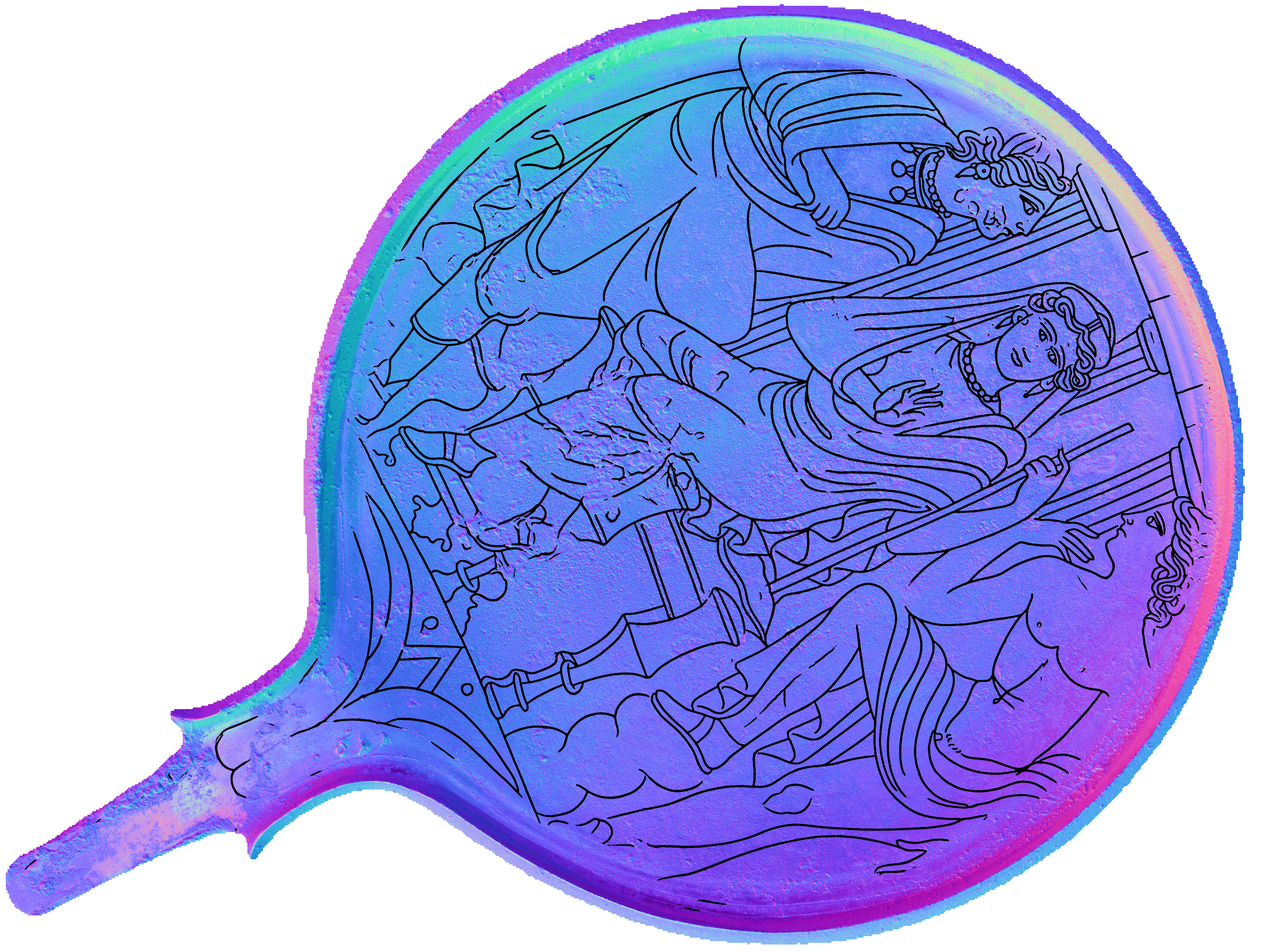}
   \caption{Showcasing performance qualitatively using mirror ``ANSA-VI-3385''.}
\end{figure}

\end{document}